\documentclass[10pt,conference]{IEEEtran}
\IEEEoverridecommandlockouts

\usepackage{cite}
\usepackage{amsmath,amssymb,amsfonts}
\usepackage{algorithmic}
\usepackage{graphicx}
\usepackage{textcomp}
\usepackage{xcolor}
\usepackage{booktabs}
\usepackage{float}
\usepackage{multirow}
\usepackage{hyperref}
\def\BibTeX{{\rm B\kern-.05em{\sc i\kern-.025em b}\kern-.08em
    T\kern-.1667em\lower.7ex\hbox{E}\kern-.125emX}}

\usepackage{pbalance}

\usepackage{comment}

\begin{document}

\title{Evaluating Tabular Representation Learning for Network Intrusion Detection}

\author{%
\IEEEauthorblockN{Muhammad Usman Butt\IEEEauthorrefmark{1}, Andreas Hotho\IEEEauthorrefmark{2}, and Daniel Schlör\IEEEauthorrefmark{2}}
\IEEEauthorblockA{\textit{Data Science Chair} \\
\textit{Center for Artificial Intelligence and Data Science} \\
\textit{Julius-Maximilians-Universität Würzburg} \\
Würzburg, Germany \\
\IEEEauthorrefmark{1}m.usman.butt@outlook.com, \\
\IEEEauthorrefmark{2}\{hotho, schloer\}@informatik.uni-wuerzburg.de}
}

\maketitle

\begin{abstract}
Classic Network Intrusion Detection Systems (NIDS) often rely on manual feature engineering to extract meaningful patterns from network traffic data. However, this approach requires domain expertise and runs counter to the widely adopted principle of modern machine learning and neural networks: that models themselves should learn meaningful representations directly from data. We investigate whether tabular representation learning techniques can improve intrusion detection performance by automatically learning robust feature representations for NetFlow data. This paper presents a systematic evaluation of state-of-the-art representation learning methods on benchmark NetFlow datasets, comparing against traditional autoencoders and end-to-end transformer baselines. We evaluate learned representations using both supervised classifiers and unsupervised anomaly detectors, with comprehensive hyperparameter exploration for each combination. 
Our results reveal strong dataset-model dependency, with no single approach consistently dominating across all scenarios. For supervised classification, TabICL achieves the best performance on CIDDS, while autoencoders follow closely and tie with end-to-end transformer models for the best average rank across datasets. Supervised approaches substantially outperform unsupervised anomaly detection methods, where no single combination consistently dominates as optimal choices depend on the dataset.
Cross-dataset transfer experiments demonstrate that learned representations can generalize across network environments with appropriate method and classifier selection. However, transfer performance varies substantially depending on the source-target dataset combination, indicating sensitivity to distributional differences between network environments. 

\end{abstract}


\section{Introduction}

Network Intrusion Detection Systems (NIDS) are essential components of modern cybersecurity infrastructure, protecting networks from unauthorized access and malicious activities. 
Traditional NIDS approaches rely heavily on manually engineered features extracted from network traffic data, a process that is labor-intensive and reflects the limitations of conventional learning algorithms: their inability to automatically extract and organize discriminative information from raw data~\cite{bengio2014representationlearningreviewnew}.
This limitation becomes particularly critical when processing NetFlow data, a structured, tabular representation of network communications that captures attributes such as source and destination addresses, ports, protocols, packet counts, and flow durations~\cite{ciscoSystems2018}.

NetFlow-based intrusion detection faces three fundamental challenges. First, the heterogeneity of features in terms of their type (e.g. continuous, categorical, ordinal) and magnitude (e.g. single vs. thousands of bytes) create a complex feature space where traditional machine learning models struggle to identify discriminative patterns~\cite{cao2022encode}. Second, severe class imbalance dominates real-world traffic, where benign flows make up most of the traffic while rare but critical attacks remain underrepresented~\cite{imbalanceddata_set}. Third, categorical features with high cardinality (e.g., IPs, ports) complicate traditional encoding schemes and limit model generalization~\cite{ring2019flow,liang2025efficientrepresentationshighcardinalitycategorical}.
Representation learning has revolutionized computer vision and natural language processing by automatically discovering meaningful data transformations~\cite{bengio2014representationlearningreviewnew}. Recent advances have adapted these techniques to tabular data, offering promise for network security applications. Methods like TabICL~\cite{qu2025tabicltabularfoundationmodel}, SCARF~\cite{bahri2022scarfselfsupervisedcontrastivelearning}, and class-conditioned contrastive learning~\cite{tabulardatacontrastivelearning} demonstrated improvements on general tabular datasets, but their effectiveness for network intrusion detection remains underexplored.

This paper presents a systematic evaluation of modern tabular representation learning techniques for NetFlow-based intrusion detection. We investigate three key questions: (1) Which representation learning methods achieve the most reliable detection performance when evaluated with both supervised classifiers and unsupervised anomaly detectors? (2) How well do representations learned on one network environment generalize to different datasets? (3) What fundamental limitations constrain the deployment of these methods in practice?

Our contributions are as follows:
\begin{itemize}
\item We conduct the first comprehensive comparison of tabular representation learning methods for network intrusion detection on three benchmark NetFlow datasets.
\item We systematically evaluate learned representations with both supervised classifiers and unsupervised anomaly detectors, revealing that method effectiveness is highly dependent on the specific dataset-model combination.
\item We demonstrate that learned representations can transfer across network environments, though with substantial variance across source-target combinations. We also identify systematic failures in detecting rare attacks, with critical implications for practical deployment.
\end{itemize}

\section{Background and Related Work}

\subsection{NetFlow and Network Intrusion Detection}

NetFlow is a network protocol developed by Cisco for collecting IP traffic information~\cite{ciscoSystems2018}. NetFlow records capture aggregated metadata about network traffic flows, including source and destination IP addresses, ports, protocols, packet and byte counts, flow duration, and TCP flags~\cite{cidds}. 

NetFlow-based NIDS analyze flow records to detect malicious activities e.g., port scans, denial-of-service attacks, brute force attempts, and botnet communication~\cite{mondragon2025advanced}. Early signature-based systems used predefined rules but struggled with zero-day attacks and attack variations while machine learning approaches addressed this limitation by learning patterns from flow data~\cite{abdulganiyu2023systematic}.

Traditional machine-learning-based NetFlow NIDS often employ shallow machine learning algorithms such as Random Forests, Support Vector Machines, and gradient boosting methods (XGBoost, LightGBM) applied to manually engineered features~\cite{abdulganiyu2023systematic}. Deep learning approaches have extended these methods with fully connected neural networks, convolutional neural networks for sequential flow analysis, and deep belief networks for hierarchical feature extraction~\cite{gamage2020deep}. More recently, graph neural networks have modeled network topology alongside flow attributes by representing hosts as nodes and flows as edges~\cite{caville2022anomal}. 
While graph-based methods introduce additional modeling assumptions by explicitly constructing network topologies, NetFlow data is fundamentally recorded and stored as flow-level tabular records. Despite this natural representation and the recent advances in tabular representation learning~\cite{jiang2025tabularsurvey}, the application of general-purpose tabular representation methods to NetFlow-based NIDS remains largely unexplored. 
Motivated by this gap, we focus on tabular representation learning methods that operate directly on flow-level records without requiring graph construction or topology modeling, examining whether these more general approaches can learn representations that transfer effectively across different network environments and attack scenarios.

\subsection{Representation Learning for Tabular Data}

Representation learning is a subfield of machine learning focused on automatically discovering meaningful data transformations that capture its essential structure and semantics~\cite{bengio2014representationlearningreviewnew}. The form of these representations often varies depending on the specific application domain. Historically, representation learning has shown significant success in areas such as Natural Language Processing and Computer Vision~\cite{doersch2015unsupervised}. More recently, progress has been made in adapting representation learning techniques to better suit tabular data~\cite{jiang2025tabularsurvey} addressing their unique challenges such as heterogeneous feature types, domain-specific relationships, and high-cardinality categorical variables that also exist in NetFlow data.

Recent work has developed specialized methods for tabular representation learning. TabTransformer~\cite{huang2020tabtransformertabulardatamodeling}, built on self-attention, transforms the embeddings of categorical features into robust contextual embeddings to achieve higher prediction accuracy but can struggle with overfitting in high-cardinality domains like network data. SCARF (Self-supervised Contrastive Augmented Representation Fine-tuning) is a self-supervised learning framework designed to learn robust data representations, particularly for tabular data where labeled data is scarce~\cite{bahri2022scarfselfsupervisedcontrastivelearning}. In SCARF, self-supervised pretraining is carried out by generating paired views of each training example through marginal corruption. 

TabICL (Tabular In-Context Learning)~\cite{qu2025tabicltabularfoundationmodel} is a scalable foundation model for tabular classification, designed to efficiently handle large tabular datasets~\cite{qu2025tabicltabularfoundationmodel}. It leverages a distribution-aware feature 
embedding to represent different types of features in a unified manner, promising enhanced transferability across diverse datasets.
In TabICL, the embedding module encodes tabular data into dense vector representations capturing statistical regularities within each column and row-wise dependencies across columns via attention.

Class-Conditioned and Feature-Correlation Based Augmentation (CC)~\cite{tabulardatacontrastivelearning} improves contrastive learning for tabular data by enhancing the data augmentation strategy, which is more challenging than in other domains due to the lack of obvious semantically consistent augmentations for tables~\cite{chen2020simpleframeworkcontrastivelearningvisual}. In contrastive learning, models learn representations by bringing together similar views (augmentations of the same record) while pushing apart dissimilar ones~\cite{contrastiveLearning}.

CC addresses the augmentation with two strategies: (1) class-conditioned corruption, which replaces feature values only with those from the same class to preserve label consistency, and (2) correlation-based masking, which selects features to corrupt based on low pairwise correlation, thus maintaining realistic feature dependencies.

For network intrusion detection, several prior works have leveraged deep learning to learn representations from NetFlow and similar data. Approaches include using Autoencoders~\cite{lopes2022effective} to obtain compressed representations for anomaly detection, modeling explicit and implicit feature correlations without attention mechanisms~\cite{wang2022representation}, and framing flow tables as images to enable convolutional neural network architectures~\cite{zhang2019multiple}. However, these earlier methods do not specifically address the unique challenges of modeling tabular network data with heterogeneous features and high-cardinality categorical variables, nor do they leverage advances from the latest state-of-the-art tabular representation learning literature. Our work addresses this gap by systematically benchmarking these methods against both traditional and end-to-end baselines across diverse network environments.

\section{Methodology}

\subsection{Datasets and Attack Landscape}

We evaluate on three benchmark NetFlow datasets that provide diverse network traffic patterns and attack types. 

CIDDS-001 (CIDDS)~\cite{cidds} contains 31.29 million flows captured over four weeks from a synthetic network environment, combining simulated benign traffic with authentic attack traffic. It includes DoS, port scanning, and brute force attacks, with flows labeled as Benign, Attacker, Victim based on 14 core attributes including IP addresses, ports, protocol, timestamps, flow statistics, and attack metadata. 
For our experiments, we utilize the portion of the CIDDS dataset collected in the OpenStack environment, omitting data from the external server due to known labeling quality issues. Additionally, for the binary classification experiments, we treat both ``attacker'' and ``victim'' labeled flows as attack traffic, as this approach captures bidirectional attack-induced traffic in the flow records.

NF-UNSW-NB15-v2~\cite{sarhan2022towards} comprises 2.39 million flows with 43 features, containing nine attack categories including Fuzzers, Analysis, Backdoor, DoS, Exploits, Generic, Reconnaissance, Shellcode, and Worms.

NF-CSE-CIC-IDS2018-v2~\cite{sarhan2022towards} consists of 18.89 million flows with 43 features, covering diverse attacks: DDoS (1.39M samples), DoS (484K), Botnet (143K), Brute Force (121K), Infiltration (116K), and Web Attacks (3.5K) in a total of 15 sub-categories.

All datasets exhibit severe class imbalance (benign traffic: 88-96\%), reflecting real-world operational conditions. Due to computational constraints, we employ stratified random subsampling to create tractable subsets while preserving class distributions and attack diversity. We repeat each experiment with five different random seeds to verify that all non-deterministic effects have minimal influence.

\subsection{Experimental Setup}
We evaluate three tabular representation learning methods: TabICL~\cite{qu2025tabicltabularfoundationmodel}, a transformer-based foundation model with distribution-aware embeddings; SCARF~\cite{bahri2022scarfselfsupervisedcontrastivelearning}, employing self-supervised contrastive learning via marginal feature corruption; and Class-Conditioned and Feature-Correlation Based Augmentation (CC)~\cite{tabulardatacontrastivelearning}, using class-aware augmentation for contrastive learning and compare those with representations learned from a simple Autoencoder as baseline.
We evaluate the decoupled tabular representation learning approach where we first learn a representation of the data and then use that representation to train a downstream classifier and compare it to a BERT-based~\cite{devlin2019bertpretrainingdeepbidirectional} end-to-end transformer baseline.

Learned representations are evaluated using supervised classifiers (Logistic Regression, SVM, Random Forest, XGBoost) and unsupervised anomaly detectors (Isolation Forest, One-Class SVM, KNN, PCA-based detection). Multiple hyperparameter configurations are explored for each model via grid search on validation sets. We report the results from the test set based on the best-performing configuration on the validation set according to AUC-PR for each method.

Data is split using stratified random sampling (60\%~/~20\%~/~20\% for train~/~validation~/~test) to ensure class balance is preserved across all splits. For all experiments and model / dataset combinations, we ensure that the underlying data sample is consistent by seed as our aim is to provide a systematic and fair comparison of representation learning methods under consistent experimental conditions, not to establish a new state-of-the-art by numbers.

For all experiments, we report AUC-ROC and AUC-PR. For supervised settings, we also report Precision, Recall, and F1-Score. Due to the severe class imbalance present in these datasets, we focus on AUC-ROC, AUC-PR, and F1-Score, which are more robust indicators of model performance in this context, rather than Accuracy, even though the latter is the most commonly used metric in prior literature. For multi-class experiments, we compute per-class metrics and macro-averaged F1 to assess performance across all attack types.

For our transferability experiments, we first train each representation learning method on one source dataset, then apply the learned encoder to generate representations for the remaining two (unseen) target datasets without any further adaptation or fine-tuning. We then train downstream classifiers on these new representations and measure detection performance. This procedure evaluates whether the representations encode general, transferable network traffic features or are overly fitted to the source dataset characteristics by testing how well the learned features generalize to different network environments and attack profiles.

\section{Experimental Results}

\subsection{Detection Performance}

We first evaluate detection performance on the CIDDS dataset as our primary detailed example, then summarize results across all three datasets.

Table~\ref{tab:cidds_results} presents the best-performing configurations for each representation method on CIDDS. TabICL combined with Random Forest achieves F1=0.883 (AUC-ROC=0.979, AUC-PR=0.945), closely matched by Autoencoder+RF at F1=0.882. These two methods exhibit different precision-recall tradeoffs: TabICL achieves higher precision (0.901) with moderate recall (0.867), while Autoencoder prioritizes recall (0.984) with lower precision (0.799), suggesting Autoencoder adopts a more aggressive detection strategy that may produce more false positives but catches more attacks.

Among other supervised methods, XGBoost with SCARF (F1=0.745) and Logistic Regression with TabICL (F1=0.728) deliver respectable performance. However, the Class-Conditioned (CC) method consistently fails across all downstream models, achieving near-zero precision despite occasionally high recall. 
We assume this collapse happens because CC representations, overwhelmed by the dominant benign class during training, fail to preserve discriminative features for the minority ``attack'' class.

\begin{table}[t]
\caption{Detection Performance on CIDDS Dataset (Best Configurations)}
\label{tab:cidds_results}
\centering
\footnotesize
\begin{tabular}{@{}llccccc@{}}
\toprule
\textbf{Method}         & \textbf{Downstream}    & \textbf{ROC}   & \textbf{PR}    & \textbf{Prec} & \textbf{Rec}  & \textbf{F1}    \\
\midrule
TabICL                  & Random Forest          & 0.979          & 0.945          & 0.901         & 0.867         & \textbf{0.883} \\
Autoencoder             & Random Forest          & 0.986          & 0.930          & 0.799         & 0.984         & 0.882          \\
SCARF                   & XGBoost                & 0.994          & 0.977          & 0.594         & 0.990         & 0.745          \\
TabICL                  & Log.\ Regression       & \textbf{0.997} & \textbf{0.991} & 0.573         & \textbf{0.998}& 0.728           \\
Autoencoder             & Log.\ Regression       & 0.980          & 0.943          & \textbf{0.990}& 0.590         & 0.739           \\
\midrule
\multicolumn{7}{l}{\textit{Baseline: End-to-End Deep Learning}} \\
BERT                    & End-to-End             & 0.986          & 0.794          & 0.798         & 0.937         & 0.864           \\
\midrule
\multicolumn{7}{l}{\textit{Unsupervised Methods (Anomaly Ranking Quality)}} \\
Autoencoder             & Isol.\ Forest          & \textbf{0.958} & \textbf{0.694} &               &               &                 \\
Autoencoder             & PCA                    & 0.926          & 0.538          &               &               &                 \\
SCARF                   & KNN                    & 0.677          & 0.251          &               &               &                 \\
SCARF                   & Isol.\ Forest          & 0.652          & 0.246          &               &               &                 \\
TabICL                  & PCA                    & 0.642          & 0.252          &               &               &                 \\
\bottomrule
\end{tabular}
\vspace{-2mm}
\begin{flushleft}
\footnotesize{Note: Precision, Recall, and F1 omitted for unsupervised methods as they require arbitrary threshold selection; AUC metrics evaluate ranking quality.}
\end{flushleft}
\vspace{-2mm}
\end{table}

Unsupervised methods achieve only moderate anomaly ranking quality compared to supervised alternatives. Autoencoder representations demonstrate the strongest unsupervised performance across different detectors, with Isolation Forest achieving AUC-ROC=0.958, AUC-PR=0.694 and PCA-based detection reaching AUC-ROC=0.926, AUC-PR=0.538. SCARF and TabICL representations yield substantially lower ranking quality (AUC-ROC$\approx$0.64-0.68, AUC-PR$\approx$0.25), demonstrating that the representations are better suited for supervised downstream classifiers than unsupervised anomaly detectors. 
Full tables with all model configurations and hyperparameters are provided  in the supplementary material~\footnote{\url{https://professor-x.de/tabrepnids}} due to space constraints.

Table~\ref{tab:summary_results} ranks all supervised method combinations across the three datasets, revealing notable dataset-model dependency. When ranking all 17 combinations within each dataset, Autoencoder+SVM and end-to-end BERT tie for best average rank (2.3), followed by TabICL+Random Forest (3.3). According to this rank-based evaluation, no single representation-classifier combination dominates universally.

Performance patterns vary substantially across datasets: TabICL+RF achieves highest F1 on CIDDS (0.883) and CSE-CIC (0.850) but only ranks 5th on UNSW (F1=0.700). Autoencoder excels on UNSW (F1=0.801 with RF, 0.801 with SVM) but struggles on CSE-CIC (F1=0.388 with RF). BERT achieves competitive performance on UNSW (F1=0.834) and on CSE-CIC (F1=0.883) but requires more computational resources and loses the modularity benefits of learned representations that can be reused across multiple tasks.

\begin{table}[t]
\caption{Top-10 Combinations by Average Rank: Supervised (F1)}
\label{tab:summary_results}
\centering
\footnotesize
\begin{tabular}{@{}lccccc@{}}
\toprule
\textbf{Representation} & \textbf{Classifier} & \textbf{CIDDS} & \textbf{UNSW} & \textbf{CSE} & \textbf{Avg.} \\
\midrule
Autoencoder & SVM & 1 & 2 & 4 & \textbf{2.3} \\
BERT & (End-to-End) & 4 & 1 & 2 & \textbf{2.3} \\
TabICL & Random Forest & 2 & 5 & 3 & \textbf{3.3} \\
Autoencoder & Random Forest & 3 & 3 & 9 & 5.0 \\
TabICL & Logistic Regression & 8 & 6 & 7 & 7.0 \\
SCARF & Random Forest & 9 & 9 & 8 & 8.7 \\
Autoencoder & XGBoost & 5 & 4 & 17 & 8.7 \\
SCARF & XGBoost & 6 & 10 & 12 & 9.3 \\
CC & SVM & 17 & 13 & 1 & 10.3 \\
CC & XGBoost & 12 & 15 & 5 & 10.7 \\
\bottomrule
\end{tabular}
\end{table}

\begin{table}[t]
\caption{Top-10 Combinations by Average Rank: Supervised (AUC-ROC)}
\label{tab:summary_results_auroc}
\centering
\footnotesize
\begin{tabular}{@{}lccccc@{}}
\toprule
\textbf{Representation} & \textbf{Classifier} & \textbf{CIDDS} & \textbf{UNSW} & \textbf{CSE} & \textbf{Avg.} \\
\midrule
BERT & (End-to-End) & 5 & 1 & 1 & \textbf{2.3} \\
Autoencoder & Random Forest & 4 & 4 & 5 & \textbf{4.3} \\
TabICL & Random Forest & 7 & 5 & 2 & \textbf{4.7} \\
CC & XGBoost & 8 & 3 & 6 & \textbf{5.7} \\
Autoencoder & Logistic Regression & 6 & 6 & 7 & \textbf{6.3} \\
TabICL & SVM & 3 & 2 & 16 & 7.0 \\
CC & Logistic Regression & 10 & 8 & 4 & 7.3 \\
Autoencoder & SVM & 9 & 10 & 3 & 7.3 \\
TabICL & Logistic Regression & 1 & 13 & 9 & 7.7 \\
TabICL & XGBoost & 11 & 9 & 8 & 9.3 \\
\bottomrule
\end{tabular}
\end{table}

When evaluating by AUC-ROC instead of F1-Score (Table~\ref{tab:summary_results_auroc}), BERT maintains its strong performance with average rank 2.3 (matching its F1-based rank). TabICL+RF (4.7) and Autoencoder+RF (4.3) also maintain strong and consistent performance across both metrics, with Autoencoder+RF showing slightly better ranking quality by AUC-ROC (4.3 vs.\ 5.0 by F1), while TabICL+RF performs slightly better by F1 (3.3 vs.\ 4.7 by AUC-ROC). Notably, some methods show large discrepancies between ranking metrics: TabICL+Logistic Regression achieves the best AUC-ROC on CIDDS (rank 1) but only ranks 8th by F1, while TabICL+SVM ranks 3rd on CIDDS by AUC-ROC but collapses to 16th on F1. 
This demonstrates that high ranking quality (AUC-ROC) may not always correspond to high classification performance (F1), especially in the presence of severe class imbalance.

\begin{table}[t]
\caption{Top-10 Combinations by Average Rank: Unsupervised (AUC-ROC)}
\label{tab:unsupervised_auroc}
\centering
\footnotesize
\begin{tabular}{@{}lccccc@{}}
\toprule
\textbf{Representation} & \textbf{Detector} & \textbf{CIDDS} & \textbf{UNSW} & \textbf{CSE} & \textbf{Avg.} \\
\midrule
Autoencoder & Isol.\ Forest & 1 & 4 & 4 & \textbf{3.0} \\
Autoencoder & PCA & 2 & 3 & 8 & \textbf{4.3} \\
TabICL & PCA & 5 & 1 & 9 & \textbf{5.0} \\
TabICL & Isol.\ Forest & 8 & 2 & 6 & \textbf{5.3} \\
SCARF & KNN & 3 & 11 & 3 & \textbf{5.7} \\
SCARF & Isol.\ Forest & 4 & 13 & 5 & \textbf{7.3} \\
TabICL & OCSVM & 16 & 8 & 2 & \textbf{8.7} \\
CC & PCA & 15 & 10 & 1 & \textbf{8.7} \\
CC & KNN & 9 & 6 & 15 & 10.0 \\
TabICL & KNN & 12 & 5 & 14 & 10.3 \\
\bottomrule
\end{tabular}
\end{table}

For unsupervised anomaly detection (Table~\ref{tab:unsupervised_auroc}), Autoencoder paired with Isolation Forest achieves the strongest and most consistent ranking quality across all datasets (average rank 3.0), followed by Autoencoder+PCA (4.3). TabICL shows dataset-dependent performance: excelling with PCA on UNSW (rank 1) and with Isolation Forest on UNSW (rank 2), but struggling on other dataset-detector combinations (e.g., TabICL+OCSVM ranks 16th on CIDDS despite ranking 2nd on CSE-CIC). Class-Conditioned methods remain highly unstable, achieving competitive ranks on some datasets (CC+PCA: rank 1 on CSE-CIC) but collapsing on others (rank 10/15 on UNSW/CIDDS). Autoencoder representations provide the most reliable foundation for unsupervised anomaly detection across the datasets.

\clearpage
\subsection{Attack Category Coverage Analysis}

Beyond aggregate metrics, we analyze per-attack-type detection effectiveness to determine which threat categories are reliably detected by each method. The goal of this analysis is to assess whether methods that achieve strong overall results are also capable of detecting infrequent but operationally significant threats, or if critical blind spots remain in practical deployments. The full heatmaps with the per-class precision and recall scores are provided in the supplementary material.

On \textbf{CIDDS}'s three-class problem (Benign, Attack, Victim), all methods achieve near-perfect detection of Benign traffic (Precision $\geq$0.9). For the Attack class, only Logistic Regression with SCARF and TabICL, and Random Forest with TabICL show meaningful detection, while most models completely omit this class. The Victim class shows better coverage with TabICL, XGBoost, and Random Forest achieving high precision and recall.
The \textbf{UNSW-NB15} dataset (Benign plus 9 attack types) reveals more nuanced patterns. TabICL demonstrates the broadest attack coverage, detecting Exploits, Fuzzers, Generic, Reconnaissance, and Shellcode with moderate-to-good precision (0.6-0.9) and recall (0.3-0.9). However, rare classes like Backdoor, Worms, and DoS remain undetectable across all methods. 
On \textbf{CSE-CIC} with 15 attack categories, TabICL with XGBoost, SVM and Logistic Regression achieve high precision ($\geq$0.9) for frequent attacks, but detection collapses for rare categories including Bot, multiple Brute-Force variants, Infiltration, and SQL Injection. Across all datasets, CC shows similar patterns but with lower precision, while SCARF fails systematically.

To quantify balanced performance across all classes equally (rather than being dominated by the majority class), we compute Macro F1 scores by averaging per-class F1 scores. On CIDDS, TabICL achieves the highest Macro F1 (0.488), closely followed by SCARF (0.408) leaving CC (0.323) behind. On UNSW, TabICL achieves the highest Macro F1 (0.291), notably outperforming CC (0.204) and SCARF (0.085). On CSE-CIC, CC outperforms TabICL (Macro F1=0.286 vs 0.196), while SCARF (0.082) again struggles. These modest absolute values reflect the fundamental challenge: all representation learning methods bias toward frequent classes during training, systematically underrepresenting threats with low sample counts.

While TabICL provides the best balanced coverage, no method reliably detects all attack types. The systematic failure on rare attacks reveals a fundamental limitation of current representation learning approaches: these methods excel at learning representations of frequent classes but underrepresent subtle feature interactions characterizing infrequent attacks, as optimization naturally prioritizes dominant classes contributing most to the loss. This suggests that future research should explore tabular representation learning techniques explicitly designed for highly imbalanced data, such as class-aware sampling strategies or contrastive objectives emphasizing rare-class boundaries.

\subsection{Cross-Dataset Evaluation}
To evaluate the robustness and generalizability of the representation learning approaches, we conduct cross-dataset experiments  training representation models on one dataset and applying them to generate representations for the other two. This evaluates how effectively learned representations transfer across datasets with different characteristics. Below we summarize the key results while full tables are in the supplementary material.

\begin{figure}[t]
\centering
\includegraphics[width=\columnwidth]{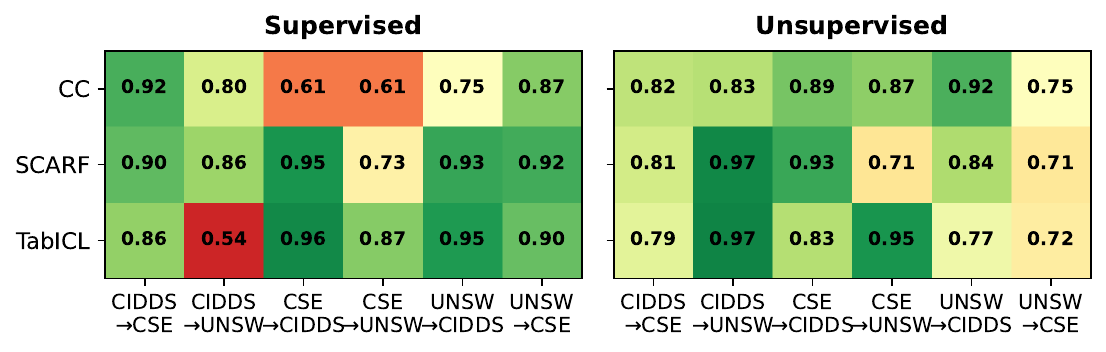}
\caption{Cross-dataset transfer learning performance (AUC-ROC) for supervised and unsupervised methods. Rows show representation learning methods (CC, SCARF, TabICL), columns show transfer paths (source$\rightarrow$target).}
\label{fig:transfer_heatmaps}
\end{figure}

Cross-dataset transfer experiments reveal that learned representations can generalize across network environments, though performance varies considerably depending on the source-target combination and learning paradigm. Figure~\ref{fig:transfer_heatmaps} visualizes the transfer performance across all method-source-target combinations.

In supervised learning, TabICL transferring from CSE-CIC to CIDDS achieves strong performance with SVM (AUC-ROC=0.96, AUC-PR=0.92), while SCARF on the same path reaches comparable results (AUC-ROC=0.95, AUC-PR=0.80). Transfers involving UNSW-NB15 as either source or target generally maintain AUC-ROC values above 0.85, with TabICL and SCARF outperforming CC across most configurations. AUC-PR values frequently exceed 0.60 in supervised settings.

Unsupervised methods show comparable transfer capability. TabICL trained on CIDDS and evaluated on UNSW-NB15 achieves high separation quality using Isolation Forest (AUC-ROC=0.97, AUC-PR=0.50), with SCARF demonstrating similarly strong transfer performance. Unsupervised AUC-PR values typically range from 0.40 to 0.50, lower than supervised counterparts (0.60-0.90). Isolation Forest serves as the dominant downstream classifier for unsupervised representations, selected in 83\% of experiments.

Transfer direction also affects performance. Transfers to CIDDS achieve the highest combined AUC-ROC and AUC-PR, particularly from CSE-CIC and UNSW-NB15 sources (supervised AUC-ROC$>$0.93). Transfers to UNSW-NB15 (supervised) show notably lower AUC-ROC (often 0.54-0.85)  showing that representations trained on other datasets do not pick up the indicative characteristics for UNSW-NB15 well. For the unsupervised models, the performance does not decline when targeting UNSW-NB15. In fact, Isolation Forest with SCARF yields the best transfer result from CIDDS to UNSW-NB15 (AUC-ROC=0.97). This suggests that deviation within the latent space during transfer is helpful for anomaly-detection-based models while harming supervised models.

Taking a broader comparative view, the results reveal notable differences between the representation learning methods. For supervised transfer, TabICL achieves the highest average performance (mean AUC-ROC=0.90), closely followed by SCARF (mean AUC-ROC=0.88), while CC shows more limited capability (mean AUC-ROC=0.76). For unsupervised transfer, all three methods perform comparably (mean AUC-ROC: CC=0.85, TabICL=0.84, SCARF=0.83), suggesting that the choice of representation learning method is more relevant for supervised than for unsupervised tasks.

The best transfer configurations reach excellent performance (AUC-ROC$>$0.95, AUC-PR$>$0.90), suggesting that pre-trained models could be leveraged from similar network environments with appropriate method and classifier selection. However, the variance in transfer performance (AUC-ROC range: 0.54-0.97) indicates the need for validating transferred models on local data before deployment. As a future direction, it would be valuable to investigate in-depth how much data from the target (transferred-to) domain, possibly combined with small-scale finetuning, is needed to further improve the transferability of embedding models for network intrusion detection.
\section{Discussion and Security Implications}

\subsection{Key Observations}
Our experimental results reveal several important patterns about representation learning effectiveness for network intrusion detection:

No single representation method consistently dominates across all datasets and downstream models. TabICL achieves best overall performance for supervised classification when paired with Random Forest (F1) or Logistic Regression (ROC, PR, Recall), closely followed by Autoencoder (Precision). 
However, Autoencoder+SVM achieves the best average rank (2.3) across all datasets, tied with end-to-end BERT. For unsupervised anomaly detection, the best method depends on the specific dataset characteristics, with no clear universal winner.

Supervised classifiers substantially outperform unsupervised anomaly detectors across all representation methods. While Autoencoder+Isolation Forest achieves moderate ranking quality, this falls short of supervised alternatives. Expectedly, supervised methods better exploit learned representations for intrusion detection for known attack types.

The BERT-based end-to-end approach achieves competitive or superior performance, but representation learning approaches offer modularity benefits since representations can be computed once and reused across multiple models.

Despite achieving competitive overall performance, all evaluated methods fundamentally struggle with extreme class imbalance. Rare but critical attacks (backdoors, infiltration, etc.) remain largely undetectable in our experiments, creating exploitable blind spots for  attackers employing low-volume techniques and emphasizing the problem with supervised methods in general.

Cross-dataset experiments show that representations can transfer across network environments, though performance varies substantially by source-target combination, especially for supervised models, necessitating validation on local data before deployment.

\subsection{Limitations and Future Research Directions}

This work assesses performance using benchmark datasets but does not evaluate robustness to adversarial perturbations specifically crafted to evade learned representations, nor does it address evaluation at real-world operational scale. These limitations are shared by much of the current research in the field and may explain the gap between results on benchmarks and outcomes observed in actual deployments. Better datasets comprising of more diverse and complex attacks and especially more realistic benign network traffic would be needed to better evaluate the robustness of models in practice.
A further methodological caveat concerns hyperparameter selection: although the representation learning and anomaly detectors are themselves label-free, we use validation-set AUC-PR to select final configurations. This deliberately focuses on the upper-bound performance achievable from each representation and minimizes confounding from hyperparameter sensitivity, but does not reflect a fully label-free deployment, where alternative selection strategies would be needed. This perspective is also a highly relevant challenge beyond NIDS in the field of anomaly detection~\cite{rochner2025we}.

Our evaluation uses stratified random splits within each dataset, which does not explicitly account for temporal distribution shift or \textit{concept drift}. Evaluating on datasets specifically designed to test for concept drift, or adopting temporal splitting strategies, would be a valuable future direction that potentially decreases the gap between supervised and unsupervised methods.

Specialized techniques to balance the class distribution, such as synthetic minority over-sampling (SMOTE)~\cite{chawla2002smote} or other data augmentation techniques~\cite{wolf2024systematic} or novel representation-learning techniques that explicitly consider class imbalance could improve detection of rare attack types without sacrificing majority class performance.

Developing transfer learning strategies that explicitly model distributional shift between network environments would enable pre-trained models to adapt efficiently to new deployments with minimal local data. The question of how much data from the target environment is needed to improve the transferability of embedding models for network intrusion detection is also an interesting future direction to investigate.

Finally, understanding which learned features contribute to detection decisions remains critical for security operations. Leveraging feature-relevance approaches~\cite{tritscher2023evaluating} in the context of tabular representations would make it possible to better understand contributions of individual features in the latent space, interpret model detections, and assess the transferability of learned representations across environments.

\section{Conclusion}
This paper presents a comprehensive evaluation of modern tabular representation learning methods (TabICL, SCARF, and Class-Conditioned contrastive learning) for NetFlow-based NIDS. Through systematic experiments across three benchmark datasets (CIDDS-001, NF-UNSW-NB15-v2, NF-CSE-CIC-IDS2018-v2) evaluating all combinations of representation methods with supervised classifiers and unsupervised anomaly detectors, we provide insights into when and how representation learning can improve detection performance.

Our results reveal strong dataset-model dependency, with no single approach consistently dominating across all scenarios. For supervised classification, TabICL achieves best overall performance on CIDDS, while the Autoencoder follows closely and ties for overall best average rank with BERT as end-to-end approach. Supervised approaches substantially outperform unsupervised anomaly detection methods, which achieve only moderate ranking quality. For unsupervised detection, no method consistently dominates as optimal combinations depend on the dataset.

However, our work also reveals important considerations for real-world deployment. Cross-dataset experiments show that representations can transfer across network environments. Yet performance varies substantially across source-target combinations and representation learning used, requiring validation on local data before deployment. More critically, per-attack-type analysis exposes systematic failures in detecting rare attacks. 

These findings have important implications: (1) Method selection must account for dataset characteristics and downstream model choices, as no universal solution exists. (2) Cross-environment deployment is feasible with appropriate method and classifier selection, though the variance in transfer performance necessitates validation on the target domain and further research into domain adaptation strategies. (3) Rare attack detection remains a fundamental unsolved problem, with all methods systematically failing on infrequent attack types.

\section*{Acknowledgment}
This work was supported by the Bavarian Ministry of Economic Affairs, Regional Development and Energy through the BRACE-LLM project (grant no. DIK-2407-0004 // DIK0726/03)

\bibliographystyle{IEEEtran}
\bibliography{bibliography}

\onecolumn
\section*{Appendix}
\subsection*{Dataset Class Distribution}

This section presents the class distribution statistics for the datasets used in our experiments.

\begin{table}[H]
\centering
\caption{UNSW-NB15 Dataset Class Distribution}
\label{tab:unsw_class_dist}
\begin{tabular}{@{}lrc@{}}
\toprule
\textbf{Class Name} & \textbf{Count} & \textbf{Percentage} \\
\midrule
Benign & 2,295,222 & 96.02\% \\
Exploits & 31,551 & 1.32\% \\
Fuzzers & 22,310 & 0.93\% \\
Generic & 16,560 & 0.69\% \\
Reconnaissance & 12,779 & 0.53\% \\
DoS & 5,794 & 0.24\% \\
Analysis & 2,299 & 0.10\% \\
Backdoor & 2,169 & 0.09\% \\
Shellcode & 1,427 & 0.06\% \\
Worms & 164 & 0.01\% \\
\midrule
\textbf{Total} & \textbf{2,390,275} & \textbf{100.00\%} \\
\bottomrule
\end{tabular}
\end{table}

\begin{table}[H]
\centering
\caption{CIDDS Dataset Class Distribution}
\label{tab:cidds_class_dist}
\begin{tabular}{@{}lrc@{}}
\toprule
\textbf{Class Name} & \textbf{Count} & \textbf{Percentage} \\
\midrule
Normal & 28,051,906 & 89.66\% \\
Attacker & 1,656,605 & 5.29\% \\
Victim & 1,579,422 & 5.05\% \\
\midrule
\textbf{Total} & \textbf{31,287,933} & \textbf{100.00\%} \\
\bottomrule
\end{tabular}
\end{table}

\begin{table}[H]
\centering
\caption{NF-CSE-CIC-IDS2018-v2 Dataset Class Distribution}
\label{tab:cic_class_dist}
\begin{tabular}{@{}lrc@{}}
\toprule
\textbf{Class Name} & \textbf{Count} & \textbf{Percentage} \\
\midrule
Benign & 16,635,567 & 88.05\% \\
DDOS attack-HOIC & 1,080,858 & 5.72\% \\
DoS attacks-Hulk & 432,648 & 2.29\% \\
DDoS attacks-LOIC-HTTP & 307,300 & 1.63\% \\
Bot & 143,097 & 0.76\% \\
Infilteration & 116,361 & 0.62\% \\
SSH-Bruteforce & 94,979 & 0.50\% \\
DoS attacks-GoldenEye & 27,723 & 0.15\% \\
FTP-BruteForce & 25,933 & 0.14\% \\
DoS attacks-SlowHTTPTest & 14,116 & 0.07\% \\
DoS attacks-Slowloris & 9,512 & 0.05\% \\
Brute Force -Web & 2,143 & 0.01\% \\
DDOS attack-LOIC-UDP & 2,112 & 0.01\% \\
Brute Force -XSS & 927 & 0.00\% \\
SQL Injection & 432 & 0.00\% \\
\midrule
\textbf{Total} & \textbf{18,893,708} & \textbf{100.00\%} \\
\bottomrule
\end{tabular}
\end{table}

\clearpage

\subsection*{Full Tables for All Models and Datasets}

\begin{table}[t]
\centering
\small
\caption{Best Supervised and Unsupervised Model Results on CIDDS-001} 
\label{table:cidds_supp}
\resizebox{0.99\textwidth}{!}{%
\begin{tabular}{@{}llp{9cm}ccc@{}}
\toprule
\textbf{Model} & \textbf{RL-Model} & \textbf{Configuration} & \textbf{AUCROC} & \textbf{AUCPR} & \textbf{F1-Score} \\ 
\midrule

\multicolumn{6}{l}{\textit{PCA-based Models}} \\[3pt]
PCA & SCARF & contamination=0.5, svd\_solver=auto n\_components=10  & 0.534  & 0.166  & 0.398  \\
PCA & TabICL & contamination=0.25, svd\_solver=randomized n\_components=5  & 0.642  & 0.252  & 0.237  \\
PCA & CC & contamination=0.1, svd\_solver=auto n\_components=5  & 0.369  & 0.001  & 0.000  \\
PCA & Autoencoder & contamination=0.5, svd\_solver=randomized n\_components=5  & 0.926  & 0.538  & 0.473  \\
\midrule

\multicolumn{6}{l}{\textit{One-Class SVM Models}} \\[3pt]
OCSVM & SCARF & nu=0.9, degree=2 gamma=auto  & 0.625  & 0.222  & 0.315  \\
OCSVM & TabICL & nu=0.5, degree=2 gamma=scale  & 0.276  & 0.115  & 0.174  \\
OCSVM & CC & nu=0.1, degree=2 gamma=scale  & 0.413  & 0.001  & 0.003  \\
OCSVM & Autoencoder & nu=0.1, degree=2 gamma=scale  & 0.376  & 0.168  & 0.281  \\
\midrule

\multicolumn{6}{l}{\textit{Isolation Forest Models}} \\[3pt]
IForest & SCARF & n\_estimators=200, max\_samples=50, max\_features=0.7, contamination=0.5  & 0.652  & 0.246  & 0.350  \\
IForest & TabICL & n\_estimators=200, max\_samples=50, max\_features=1.0, contamination=0.5 & 0.550  & 0.170  & 0.136   \\
IForest & CC & n\_estimators=200, max\_samples=50, max\_features=0.7, contamination=0.5  & 0.399 & 0.001  & 0.002  \\
IForest & Autoencoder & n\_estimators=200, max\_samples=50, max\_features=0.7, contamination=0.25  & 0.958 & 0.694  & 0.791  \\
\midrule

\multicolumn{6}{l}{\textit{K-Nearest Neighbors Models}} \\[3pt]
KNN & SCARF & n\_neighbors=10, algorithm=auto, metric=euclidean  & 0.677 & 0.251  & 0.198  \\
KNN & TabICL & n\_neighbors=3, algorithm=auto, metric=cosine  & 0.412 & 0.136  & 0.150  \\
KNN & CC & n\_neighbors=3, algorithm=auto, metric=manhattan  & 0.540 & 0.001  & 0.003  \\
KNN & Autoencoder & n\_neighbors=5, algorithm=brute, metric=manhattan  & 0.637 & 0.198 & 0.316  \\
\midrule

\multicolumn{6}{l}{\textit{Support Vector Machine Models}} \\[3pt]
SVM & SCARF & gamma=scale, coef0=0.5, tol=0.1, C=1.0, kernel=linear & 0.764 & 0.404  & 0.447  \\
SVM & TabICL & gamma=scale, coef0=0.0, tol=0.0001, C=0.1, kernel=linear & 0.991 & 0.970  & 0.045  \\
SVM & CC & gamma=scale, coef0=0.0, tol=0.0001, C=0.1, kernel=rbf & 0.896 & 0.094 & 0.005  \\
SVM & Autoencoder & gamma=scale, coef0=0.0, tol=0.0001, C=0.1, kernel=linear & 0.970 & 0.768 & 0.936  \\
\midrule

\multicolumn{6}{l}{\textit{Logistic Regression Models}} \\[3pt]
LR & SCARF & penalty=l2, C=0.1, class\_weight=balanced, solver=lbfgs & 0.914 & 0.665  & 0.389  \\
LR & TabICL & penalty=l2, C=0.1, class\_weight=balanced, solver=liblinear & 0.997 & 0.991  & 0.728 \\
LR & CC & penalty=none, C=0.1, class\_weight=none, solver=newton-cg & 0.956 & 0.236  & 0.059  \\
LR & Autoencoder & penalty=l2, C=10, class\_weight=none, solver=lbfgs & 0.980 & 0.943  & 0.739  \\
\midrule

\multicolumn{6}{l}{\textit{Random Forest Models}} \\[3pt]
RF & SCARF & n\_estimators=200, max\_depth=50, class\_weight=balanced, ccp\_alpha=0.01 & 0.869 & 0.625  & 0.614 \\
RF & TabICL & n\_estimators=200, max\_depth=10, class\_weight=balanced, ccp\_alpha=0.1 & 0.979 & 0.945  & 0.883 \\
RF & CC & n\_estimators=200, max\_depth=50, class\_weight=balanced, ccp\_alpha=0 & 0.928 & 0.524& 0.515\\
RF & Autoencoder & n\_estimators=200, max\_depth=10, class\_weight=balanced, ccp\_alpha=0 & 0.986 & 0.930& 0.882\\
\midrule

\multicolumn{6}{l}{\textit{XGBoost Models}} \\[3pt]
XGBoost & SCARF & lr=0.01, max\_depth=3, n\_estimators=100, reg\_alpha=0, reg\_lambda=1  & 0.994 & 0.977 & 0.745\\
XGBoost & TabICL & lr=0.1, max\_depth=5, n\_estimators=100, reg\_alpha=0, reg\_lambda=0.1  & 0.940 & 0.719 & 0.171\\
XGBoost & CC & lr=0.01, max\_depth=5, n\_estimators=1000, reg\_alpha=0.5, reg\_lambda=0.1  & 0.972 & 0.449 & 0.429\\
XGBoost & Autoencoder & lr=0.01, max\_depth=3, n\_estimators=100, reg\_alpha=0, reg\_lambda=0.1  & 0.911 & 0.733 & 0.787 \\
\midrule

\multicolumn{6}{l}{\textit{End-to-End Deep Learning Models}} \\[3pt]
BERT & End-to-End (classification head) & lr=1e-5, batch\_size=32, epochs=4 & 0.986 & 0.794 & 0.864 \\
Autoencoder & End-to-End (classification head) & latent\_dim=16 & 0.963 & 0.733 & 0.608 \\
\bottomrule

\end{tabular}
}
\end{table}


\begin{table}[H]
\centering
\small
\caption{Best Supervised and Unsupervised Model Results on NF-UNSW-NB15-v2} 
\label{table:unsw_supp}
\resizebox{1\textwidth}{!}{%
\begin{tabular}{@{}llp{9cm}ccc@{}}
\toprule
\textbf{Model} & \textbf{RL-Model} & \textbf{Configuration} & \textbf{AUCROC} & \textbf{AUCPR} & \textbf{F1-Score} \\ 
\midrule

\multicolumn{6}{l}{\textit{PCA-based Models}} \\[3pt]
PCA & SCARF & contamination=0.1, svd\_solver=randomized n\_components=15  & 0.591  & 0.151  & 0.222 \\
PCA & TabICL & contamination=0.25, svd\_solver=randomized n\_components=5  & 0.976  & 0.481  & 0.488  \\
PCA & CC & contamination=0.1, svd\_solver=auto n\_components=5  & 0.835  & 0.009  & 0.015  \\
PCA & Autoencoder & contamination=0.1, svd\_solver=auto n\_components=5  & 0.948 & 0.615  & 0.540 \\
\midrule

\multicolumn{6}{l}{\textit{One-Class SVM Models}} \\[3pt]
OCSVM & SCARF & nu=0.9, degree=2 gamma=scale  & 0.526  & 0.114  & 0.234  \\
OCSVM & TabICL & nu=0.5, degree=2 gamma=scale  & 0.856 & 0.387  & 0.175  \\
OCSVM & CC & nu=0.1, degree=2 gamma=auto  & 0.878  & 0.018  & 0.002  \\
OCSVM & Autoencoder & nu=0.1, degree=3 gamma=scale  & 0.448  & 0.185  & 0.274  \\
\midrule

\multicolumn{6}{l}{\textit{Isolation Forest Models}} \\[3pt]
IForest & SCARF & n\_estimators=200, max\_samples=100, max\_features=0.5, contamination=0.5  & 0.612  & 0.156  & 0.248  \\
IForest & TabICL & n\_estimators=200, max\_samples=100, max\_features=0.7, contamination=0.1 & 0.974  & 0.470  & 0.482   \\
IForest & CC & n\_estimators=200, max\_samples=50, max\_features=0.7, contamination=0.5  & 0.854 & 0.008  & 0.021  \\
IForest & Autoencoder & n\_estimators=200, max\_samples=50, max\_features=0.5, contamination=0.1 & 0.926  & 0.545 & 0.425  \\
\midrule

\multicolumn{6}{l}{\textit{K-Nearest Neighbors Models}} \\[3pt]
KNN & SCARF & n\_neighbors=5, algorithm=auto, metric=euclidean  & 0.674 & 0.075  & 0.050  \\
KNN & TabICL & n\_neighbors=10, algorithm=auto, metric=manhattan  & 0.917 & 0.346 & 0.229  \\
KNN & CC & n\_neighbors=3, algorithm=auto, metric=minkowski  & 0.882 & 0.028 & 0.001  \\
KNN & Autoencoder & n\_neighbors=5, algorithm=auto, metric=euclidean  & 0.642 & 0.198 & 0.300  \\
\midrule

\multicolumn{6}{l}{\textit{Support Vector Machine Models}} \\[3pt]
SVM & SCARF & gamma=scale, coef0=0.5, tol=0.1, C=1.0, kernel=linear & 0.764 & 0.404  & 0.447  \\
SVM & TabICL & gamma=scale, coef0=0.0, tol=0.0001, C=0.1, kernel=linear & 0.991 & 0.970  & 0.045  \\
SVM & CC & gamma=scale, coef0=0.0, tol=0.0001, C=0.1, kernel=rbf & 0.896 & 0.094 & 0.005  \\
SVM & Autoencoder & gamma=scale, coef0=0.0, tol=0.0001, C=10, kernel=rbf & 0.957 & 0.892 & 0.801  \\
\midrule

\multicolumn{6}{l}{\textit{Logistic Regression Models}} \\[3pt]
LR & SCARF & penalty=l2, C=0.1, class\_weight=none, solver=lbfgs & 0.873 & 0.304  & 0.072  \\
LR & TabICL & penalty=l2, C=0.1, class\_weight=none, solver=newton-cg & 0.879 & 0.546  & 0.484 \\
LR & CC & penalty=l2, C=10, class\_weight=none, solver=liblinear & 0.973 & 0.681  & 0.114  \\
LR & Autoencoder & penalty=l2, C=0.1, class\_weight=balanced, solver=sag & 0.979 & 0.860  & 0.001  \\
\midrule

\multicolumn{6}{l}{\textit{Random Forest Models}} \\[3pt]
RF & SCARF & n\_estimators=200, max\_depth=50, class\_weight=balanced, ccp\_alpha=0.01 & 0.806 & 0.157& 0.077 \\
RF & TabICL & n\_estimators=200, max\_depth=10, class\_weight=balanced, ccp\_alpha=0.1 & 0.987 & 0.844  & 0.700 \\
RF & CC & n\_estimators=10, max\_depth=10, class\_weight=None, ccp\_alpha=0 & 0.899 & 0.427& 0.004\\
RF & Autoencoder & n\_estimators=100, max\_depth=10, class\_weight=balanced, ccp\_alpha=0.01 & 0.987 & 0.934 & 0.801\\
\midrule

\multicolumn{6}{l}{\textit{XGBoost Models}} \\[3pt]
XGBoost & SCARF & lr=0.01, max\_depth=3, n\_estimators=100, reg\_alpha=0.5, reg\_lambda=1  & 0.850 & 0.317 & 0.077\\
XGBoost & TabICL & lr=0.1, max\_depth=3, n\_estimators=500, reg\_alpha=0, reg\_lambda=0.5  & 0.958 & 0.505 & 0.001\\
XGBoost & CC & lr=0.01, max\_depth=5, n\_estimators=1000, reg\_alpha=0.5, reg\_lambda=0.1  & 0.990 & 0.664 & 0.004 \\
XGBoost & Autoencoder & lr=0.01, max\_depth=7, n\_estimators=100, reg\_alpha=0, reg\_lambda=0.1  & 0.977 & 0.836 & 0.796 \\
\midrule

\multicolumn{6}{l}{\textit{End-to-End Deep Learning Models}} \\[3pt]
BERT & End-to-End (classification head) & lr=1e-5, batch\_size=32, epochs=4 & 0.996 & 0.894 & 0.834 \\
Autoencoder & End-to-End (classification head) & latent\_dim=64 & 0.991 & 0.763 & 0.722 \\
\bottomrule

\end{tabular}
}
\end{table}

\clearpage


\begin{table}[H]
\centering
\small
\caption{Best Supervised and Unsupervised Model Results on NF-CSE-CIC-IDS2018} 
\label{table:cic_supp}
\resizebox{0.98\textwidth}{!}{%
\begin{tabular}{@{}llp{9cm}ccc@{}}
\toprule
\textbf{Model} & \textbf{RL-Model} & \textbf{Configuration} & \textbf{AUCROC} & \textbf{AUCPR} & \textbf{F1-Score} \\ 
\midrule

\multicolumn{6}{l}{\textit{PCA-based Models}} \\[3pt]
PCA & SCARF & contamination=0.5, svd\_solver=randomized n\_components=5  & 0.604  & 0.156  & 0.116 \\
PCA & TabICL & contamination=0.25, svd\_solver=randomized n\_components=5  & 0.591  & 0.151  & 0.221  \\
PCA & CC & contamination=0.1, svd\_solver=auto n\_components=5  & 0.835  & 0.009  & 0.015  \\
PCA & Autoencoder & contamination=0.1, svd\_solver=randomized n\_components=5  & 0.600 & 0.153  & 0.145\\
\midrule

\multicolumn{6}{l}{\textit{One-Class SVM Models}} \\[3pt]
OCSVM & SCARF & nu=0.9, degree=2 gamma=scale  & 0.527 & 0.114 & 0.234 \\
OCSVM & TabICL & nu=0.9, degree=2 gamma=scale  & 0.793 & 0.342  & 0.084  \\
OCSVM & CC & nu=0.1, degree=2 gamma=auto  & 0.492  & 0.143  & 0.201  \\
OCSVM & Autoencoder & nu=0.1, degree=2 gamma=auto  & 0.518  & 0.122  & 0.213  \\
\midrule

\multicolumn{6}{l}{\textit{Isolation Forest Models}} \\[3pt]
IForest & SCARF & n\_estimators=200, max\_samples=100, max\_features=0.1, contamination=1.0  & 0.626  & 0.154  & 0.244  \\
IForest & TabICL & n\_estimators=200, max\_samples=100, max\_features=0.5, contamination=0.5 & 0.612  & 0.156  & 0.248   \\
IForest & CC & n\_estimators=200, max\_samples=50, max\_features=0.7, contamination=0.5  & 0.562 & 0.152 & 0.231  \\
IForest & Autoencoder & n\_estimators=200, max\_samples=50, max\_features=1.0, contamination=0.1 & 0.656  & 0.168 & 0.235  \\
\midrule

\multicolumn{6}{l}{\textit{K-Nearest Neighbors Models}} \\[3pt]
KNN & SCARF & n\_neighbors=10, algorithm=auto, metric=euclidean  & 0.704 & 0.206  & 0.120  \\
KNN & TabICL & n\_neighbors=10, algorithm=auto, metric=manhattan  & 0.491 & 0.106 & 0.099 \\
KNN & CC & n\_neighbors=10, algorithm=auto, metric=manhattan  & 0.344 & 0.107 & 0.294  \\
KNN & Autoencoder & n\_neighbors=3, algorithm=brute, metric=manhattan  & 0.234 & 0.079 & 0.238  \\
\midrule

\multicolumn{6}{l}{\textit{Support Vector Machine Models}} \\[3pt]
SVM & SCARF & gamma=scale, coef0=0.5, tol=0.0001, C=1.0, kernel=poly & 0.500 & 0.118  & 0.212 \\
SVM & TabICL & gamma=scale, coef0=0.5, tol=0.0001, C=0.1, kernel=sigmoid & 0.500 & 0.119  & 0.213  \\
SVM & CC & gamma=scale, coef0=1.0, tol=0.0001, C=10, kernel=sigmoid & 0.741 & 0.094 & 0.946  \\
SVM & Autoencoder & gamma=scale, coef0=0.0, tol=0.0001, C=10, kernel=rbf & 0.957 & 0.892 & 0.801  \\
\midrule

\multicolumn{6}{l}{\textit{Logistic Regression Models}} \\[3pt]
LR & SCARF & penalty=l2, C=0.1, class\_weight=none, solver=saga & 0.642  & 0.247 & 0.209   \\
LR & TabICL & penalty=l2, C=0.1, class\_weight=balanced, solver=sag & 0.881 & 0.405  & 0.579 \\
LR & CC & penalty=l2, C=10, class\_weight=none, solver=sag & 0.946 & 0.913  & 0.215  \\
LR & Autoencoder & penalty=l2, C=0.1, class\_weight=none, solver=liblinear & 0.887 & 0.815 & 0.204  \\
\midrule

\multicolumn{6}{l}{\textit{Random Forest Models}} \\[3pt]
RF & SCARF & n\_estimators=10, max\_depth=50, class\_weight=balanced, ccp\_alpha=0.01 & 0.868 & 0.627 & 0.479 \\
RF & TabICL & n\_estimators=200, max\_depth=50, class\_weight=balanced, ccp\_alpha=0.01 & 0.971 & 0.922  & 0.850 \\
RF & CC & n\_estimators=10, max\_depth=10, class\_weight=None, ccp\_alpha=0 & 0.729 & 0.497 & 0.216\\
RF & Autoencoder & n\_estimators=100, max\_depth=10, class\_weight=balanced, ccp\_alpha=0.1 & 0.930 & 0.826 & 0.388\\
\midrule

\multicolumn{6}{l}{\textit{XGBoost Models}} \\[3pt]
XGBoost & SCARF & lr=0.01, max\_depth=3, n\_estimators=1000, reg\_alpha=0, reg\_lambda=1  & 0.763 & 0.430 & 0.214\\
XGBoost & TabICL & lr=0.01, max\_depth=3, n\_estimators=100, reg\_alpha=0, reg\_lambda=0.1  & 0.886 & 0.443 & 0.587\\
XGBoost & CC & lr=0.01, max\_depth=5, n\_estimators=1000, reg\_alpha=0.5, reg\_lambda=0.1  & 0.891 & 0.611 & 0.640 \\
XGBoost & Autoencoder & lr=0.01, max\_depth=7, n\_estimators=100, reg\_alpha=0.1, reg\_lambda=0.1  & 0.800 & 0.550 & 0.145\\
\midrule

\multicolumn{6}{l}{\textit{End-to-End Deep Learning Models}} \\[3pt]
BERT & End-to-End (classification head) & lr=2e-5, batch\_size=16, epochs=3 & 0.9997 & 0.938 & 0.883 \\
Autoencoder & End-to-End (classification head) & latent\_dim=8 & 0.975 & 0.903 & 0.876 \\
\bottomrule

\end{tabular}
}
\end{table}

\clearpage

\subsection*{Full Performance Ranking Tables}


\begin{table}[H]
\centering
\caption{F1-Score Rankings of Supervised Models Across Datasets}
\label{tab:supervised_f1_ranking}
\begin{tabular}{@{}lcccc@{}}
\toprule
\textbf{Method} & \textbf{CIDDS Rank} & \textbf{UNSW Rank} & \textbf{CSE-CIC Rank} & \textbf{Avg. Rank} \\
\midrule
BERT & 4 & 1 & 2 & 2.33 \\
SVM + Autoencoder & 1 & 2 & 4 & 2.33 \\
RF + TabICL & 2 & 5 & 3 & 3.33 \\
RF + Autoencoder & 3 & 3 & 9 & 5.00 \\
LR + TabICL & 8 & 6 & 7 & 7.00 \\
RF + SCARF & 9 & 9 & 8 & 8.67 \\
XGBoost + Autoencoder & 5 & 4 & 17 & 8.67 \\
XGBoost + SCARF & 6 & 10 & 12 & 9.33 \\
SVM + CC & 17 & 13 & 1 & 10.33 \\
SVM + SCARF & 11 & 7 & 14 & 10.67 \\
XGBoost + CC & 12 & 15 & 5 & 10.67 \\
LR + CC & 15 & 8 & 11 & 11.33 \\
RF + CC & 10 & 14 & 10 & 11.33 \\
XGBoost + TabICL & 14 & 17 & 6 & 12.33 \\
LR + Autoencoder & 7 & 16 & 16 & 13.00 \\
LR + SCARF & 13 & 11 & 15 & 13.00 \\
SVM + TabICL & 16 & 12 & 13 & 13.67 \\
\bottomrule
\end{tabular}
\end{table}


\begin{table}[H]
\centering
\caption{AUC-ROC Rankings of Supervised Models Across Datasets}
\label{tab:supervised_roc_ranking}
\begin{tabular}{@{}lcccc@{}}
\toprule
\textbf{Method} & \textbf{CIDDS Rank} & \textbf{UNSW Rank} & \textbf{CSE-CIC Rank} & \textbf{Avg. Rank} \\
\midrule
BERT & 5 & 1 & 1 & 2.33 \\
RF + Autoencoder & 4 & 4 & 5 & 4.33 \\
RF + TabICL & 7 & 5 & 2 & 4.67 \\
XGBoost + CC & 8 & 3 & 6 & 5.67 \\
LR + Autoencoder & 6 & 6 & 7 & 6.33 \\
SVM + TabICL & 3 & 2 & 16 & 7.00 \\
LR + CC & 10 & 8 & 4 & 7.33 \\
SVM + Autoencoder & 9 & 10 & 3 & 7.33 \\
LR + TabICL & 1 & 13 & 9 & 7.67 \\
XGBoost + TabICL & 11 & 9 & 8 & 9.33 \\
XGBoost + SCARF & 2 & 15 & 12 & 9.67 \\
XGBoost + Autoencoder & 14 & 7 & 11 & 10.67 \\
RF + CC & 12 & 11 & 14 & 12.33 \\
SVM + CC & 15 & 12 & 13 & 13.33 \\
LR + SCARF & 13 & 14 & 15 & 14.00 \\
RF + SCARF & 16 & 16 & 10 & 14.00 \\
SVM + SCARF & 17 & 17 & 17 & 17.00 \\
\bottomrule
\end{tabular}
\end{table}

\clearpage


\begin{table}[H]
\centering
\caption{AUC-ROC Rankings of Unsupervised Models Across Datasets}
\label{tab:unsupervised_roc_ranking}
\begin{tabular}{@{}lcccc@{}}
\toprule
\textbf{Method} & \textbf{CIDDS Rank} & \textbf{UNSW Rank} & \textbf{CSE Rank} & \textbf{Avg. Rank} \\
\midrule
IForest + Autoencoder & 1 & 4 & 4 & 3.00 \\
PCA + Autoencoder & 2 & 3 & 8 & 4.33 \\
PCA + TabICL & 5 & 1 & 9 & 5.00 \\
IForest + TabICL & 8 & 2 & 6 & 5.33 \\
KNN + SCARF & 3 & 11 & 3 & 5.67 \\
IForest + SCARF & 4 & 13 & 5 & 7.33 \\
OCSVM + TabICL & 16 & 8 & 2 & 8.67 \\
PCA + CC & 15 & 10 & 1 & 8.67 \\
KNN + CC & 9 & 6 & 15 & 10.00 \\
KNN + TabICL & 12 & 5 & 14 & 10.33 \\
OCSVM + CC & 11 & 7 & 13 & 10.33 \\
PCA + SCARF & 10 & 14 & 7 & 10.33 \\
IForest + CC & 13 & 9 & 10 & 10.67 \\
OCSVM + SCARF & 7 & 15 & 11 & 11.00 \\
KNN + Autoencoder & 6 & 12 & 16 & 11.33 \\
OCSVM + Autoencoder & 14 & 16 & 12 & 14.00 \\
\bottomrule
\end{tabular}
\end{table}

\clearpage

\subsection*{Cross-Dataset Transfer Learning Results}

This section presents the results from the cross-dataset transfer learning experiments, where models trained on one dataset are evaluated on another selecting the best model configuration (according to validation AUC-PR) for each transfer path.

\begin{table}[H]
\centering
\caption{Cross-Dataset Transfer Learning Results (Supervised)}
\label{tab:transfer_rerun_supervised}
\resizebox{0.70\textwidth}{!}{%
\begin{tabular}{llllcc}
\toprule
Method & Source & Target & Model & AUCROC & AUCPR \\
\midrule
CC & CIDDS & CSE-CIC & SVM & 0.9185 & 0.6108 \\
CC & CIDDS & UNSW-NB15 & SVM & 0.7995 & 0.5701 \\
CC & CSE-CIC & CIDDS & SVM & 0.6110 & 0.3548 \\
CC & CSE-CIC & UNSW-NB15 & SVM & 0.6107 & 0.3811 \\
CC & UNSW-NB15 & CIDDS & LR & 0.7514 & 0.4983 \\
CC & UNSW-NB15 & CSE-CIC & SVM & 0.8733 & 0.6436 \\
\midrule
SCARF & CIDDS & CSE-CIC & LR & 0.9029 & 0.7038 \\
SCARF & CIDDS & UNSW-NB15 & SVM & 0.8567 & 0.5733 \\
SCARF & CSE-CIC & CIDDS & SVM & 0.9514 & 0.7992 \\
SCARF & CSE-CIC & UNSW-NB15 & LR & 0.7267 & 0.3923 \\
SCARF & UNSW-NB15 & CIDDS & SVM & 0.9315 & 0.7741 \\
SCARF & UNSW-NB15 & CSE-CIC & LR & 0.9238 & 0.7856 \\
\midrule
TabICL & CIDDS & CSE-CIC & SVM & 0.8642 & 0.6803 \\
TabICL & CIDDS & UNSW-NB15 & LR & 0.5405 & 0.1469 \\
TabICL & CSE-CIC & CIDDS & SVM & 0.9637 & 0.9249 \\
TabICL & CSE-CIC & UNSW-NB15 & LR & 0.8741 & 0.5547 \\
TabICL & UNSW-NB15 & CIDDS & SVM & 0.9471 & 0.8983 \\
TabICL & UNSW-NB15 & CSE-CIC & LR & 0.8978 & 0.7318 \\
\bottomrule
\end{tabular}
}%
\end{table}

\begin{table}[H]
\centering
\caption{Cross-Dataset Transfer Learning Results (Unsupervised)}
\label{tab:transfer_rerun_unsupervised}
\resizebox{0.70\textwidth}{!}{%
\begin{tabular}{llllcc}
\toprule
Method & Source & Target & Model & AUCROC & AUCPR \\
\midrule
CC & CIDDS & CSE-CIC & LOF & 0.8249 & 0.5863 \\
CC & CIDDS & UNSW-NB15 & IForest & 0.8320 & 0.4473 \\
CC & CSE-CIC & CIDDS & IForest & 0.8850 & 0.5187 \\
CC & CSE-CIC & UNSW-NB15 & IForest & 0.8681 & 0.3565 \\
CC & UNSW-NB15 & CIDDS & IForest & 0.9164 & 0.7359 \\
CC & UNSW-NB15 & CSE-CIC & PCA & 0.7489 & 0.3222 \\
\midrule
SCARF & CIDDS & CSE-CIC & IForest & 0.8056 & 0.5837 \\
SCARF & CIDDS & UNSW-NB15 & IForest & 0.9658 & 0.5167 \\
SCARF & CSE-CIC & CIDDS & IForest & 0.9284 & 0.6728 \\
SCARF & CSE-CIC & UNSW-NB15 & LOF & 0.7099 & 0.4009 \\
SCARF & UNSW-NB15 & CIDDS & IForest & 0.8402 & 0.4624 \\
SCARF & UNSW-NB15 & CSE-CIC & IForest & 0.7147 & 0.2814 \\
\midrule
TabICL & CIDDS & CSE-CIC & IForest & 0.7854 & 0.4727 \\
TabICL & CIDDS & UNSW-NB15 & IForest & 0.9693 & 0.5035 \\
TabICL & CSE-CIC & CIDDS & IForest & 0.8349 & 0.5083 \\
TabICL & CSE-CIC & UNSW-NB15 & IForest & 0.9522 & 0.5139 \\
TabICL & UNSW-NB15 & CIDDS & IForest & 0.7703 & 0.3786 \\
TabICL & UNSW-NB15 & CSE-CIC & IForest & 0.7215 & 0.3124 \\
\bottomrule
\end{tabular}
}%
\end{table}
\clearpage

\subsection*{Multiclass Classification Analysis}

This section presents per-class precision and recall visualizations for multiclass classification across the three representation learning methods (SCARF, TabICL, and CC).
\\[1em]
\emph{CIDDS-001 Dataset: Multiclass Performance}

\begin{figure}[H]
    \centering
    
  \begin{minipage}{0.45\textwidth}
    \centering
    \includegraphics[width=\linewidth]{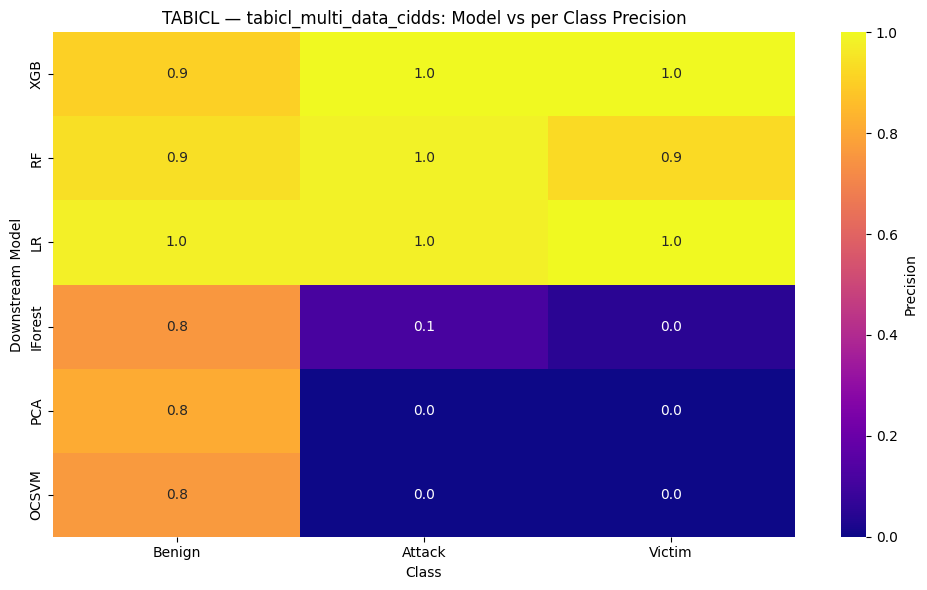}
    \caption{TabICL Precision per class on CIDDS-001}
  \end{minipage}
  \hfill
  \begin{minipage}{0.45\textwidth}
    \centering
    \includegraphics[width=\linewidth]{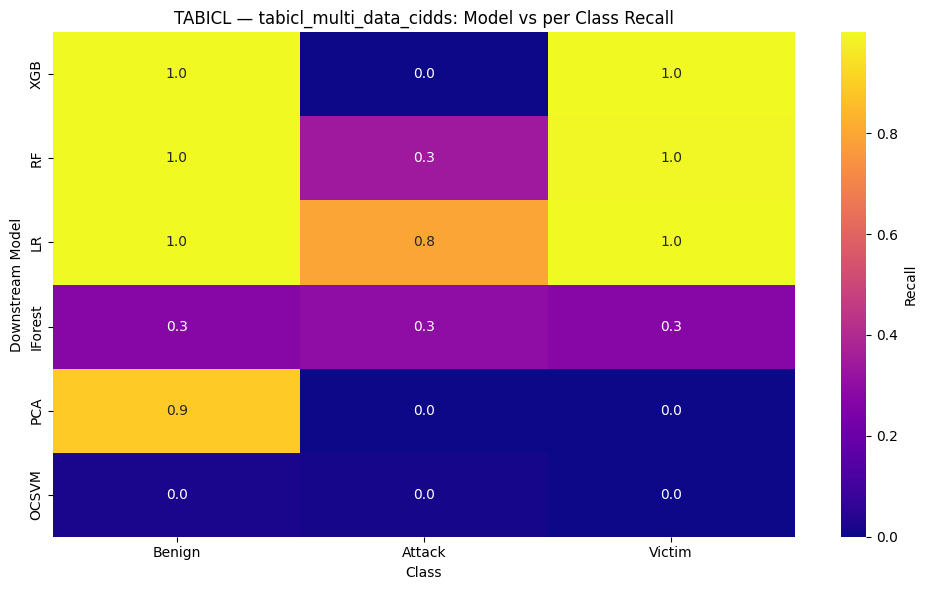}
    \caption{TabICL Recall per class on CIDDS-001}
  \end{minipage}


    \begin{minipage}{0.45\textwidth}
      \centering
      \includegraphics[width=\linewidth]{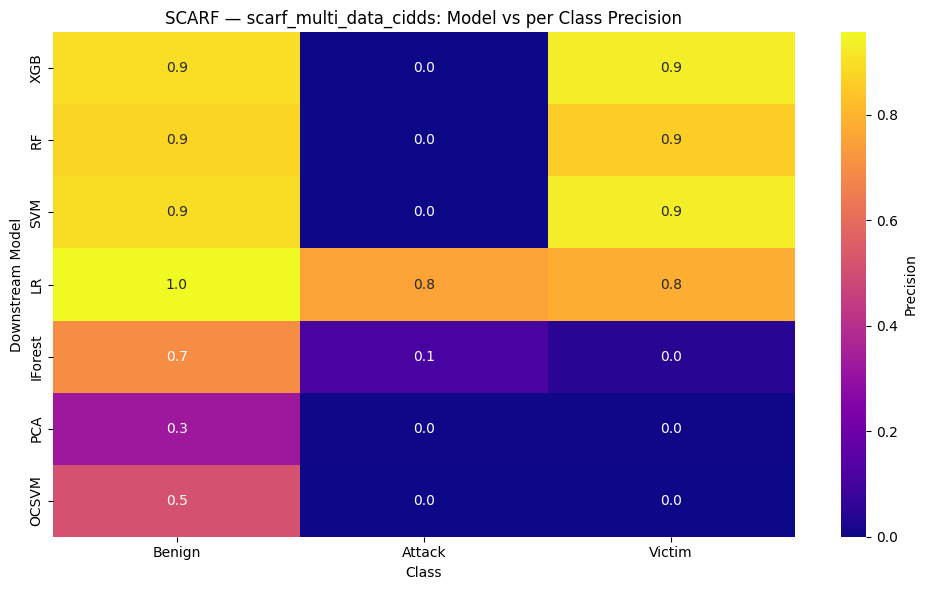}
      \caption{SCARF Precision per class on CIDDS-001}
    \end{minipage}
    \hfill
    \begin{minipage}{0.45\textwidth}
      \centering
      \includegraphics[width=\linewidth]{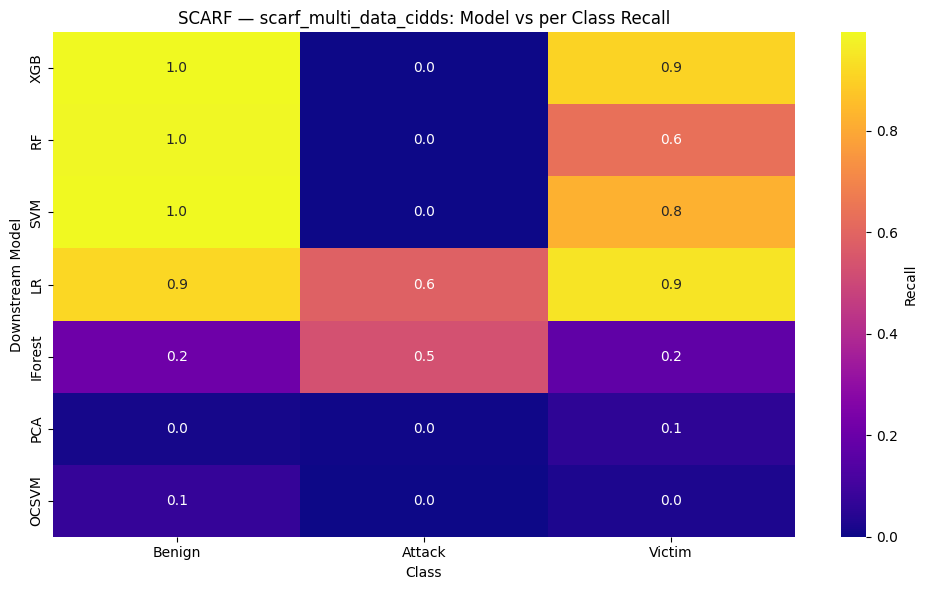}
      \caption{SCARF Recall per class on CIDDS-001}
    \end{minipage}

  
  \begin{minipage}{0.45\textwidth}
    \centering
    \includegraphics[width=\linewidth]{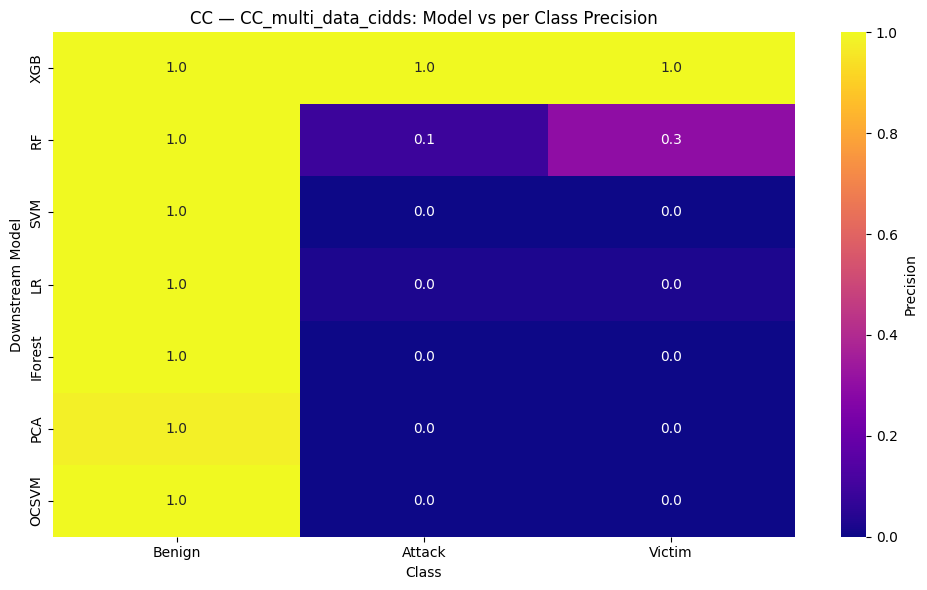}
    \caption{CC Precision per class on CIDDS-001}
  \end{minipage}
  \hfill
  \begin{minipage}{0.45\textwidth}
    \centering
    \includegraphics[width=\linewidth]{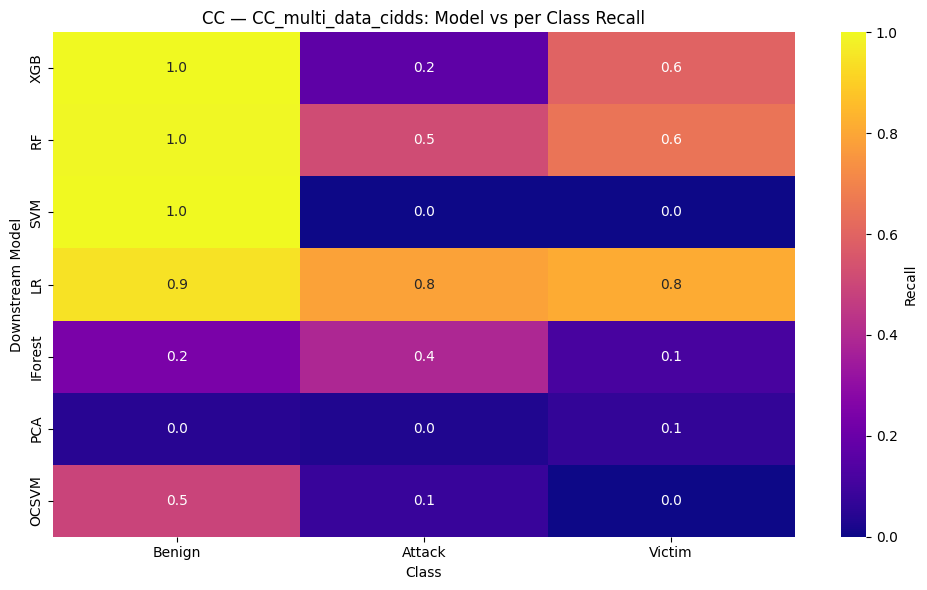}
    \caption{CC Recall per class on CIDDS-001}
  \end{minipage}
\end{figure}

\clearpage

\emph{UNSW-NB15 Dataset: Multiclass Performance}

\begin{figure}[H]
    \centering  

  \begin{minipage}{0.45\textwidth}
    \centering
    \includegraphics[width=\linewidth]{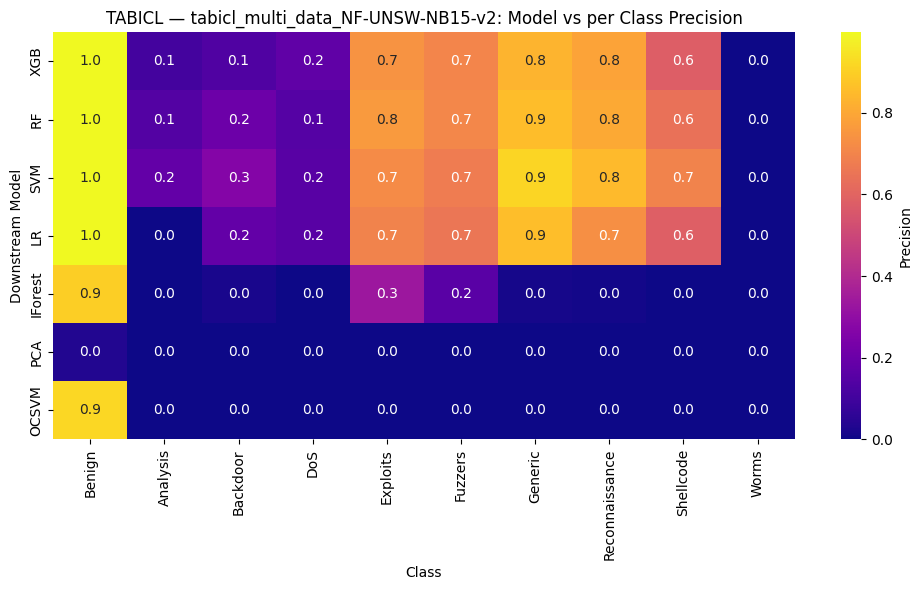}
    \caption{TabICL Precision per class on UNSW-NB15}
  \end{minipage}
  \hfill
  \begin{minipage}{0.45\textwidth}
    \centering
    \includegraphics[width=\linewidth]{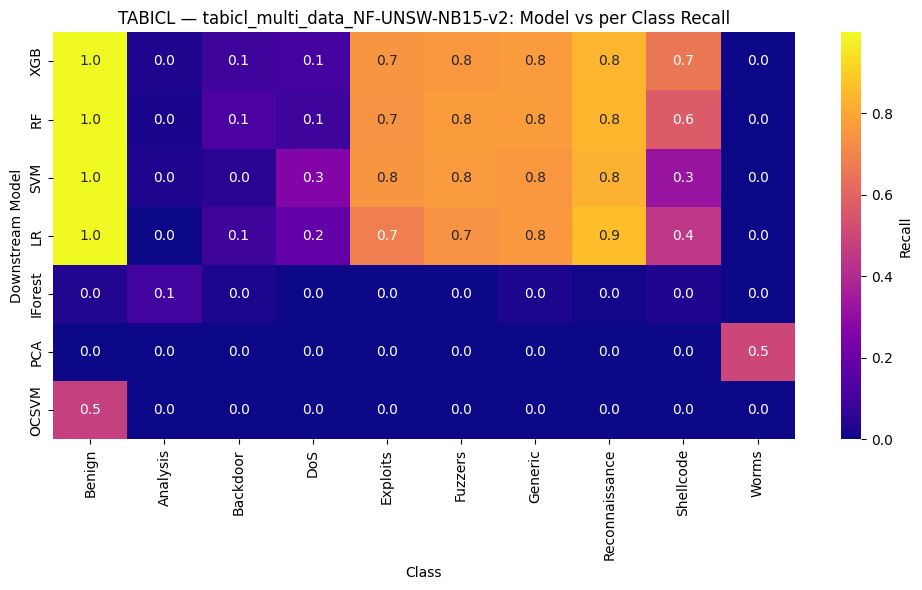}
    \caption{TabICL Recall per class on UNSW-NB15}
  \end{minipage}
  
  \vspace{0.5cm}

    \begin{minipage}{0.45\textwidth}
      \centering
      \includegraphics[width=\linewidth]{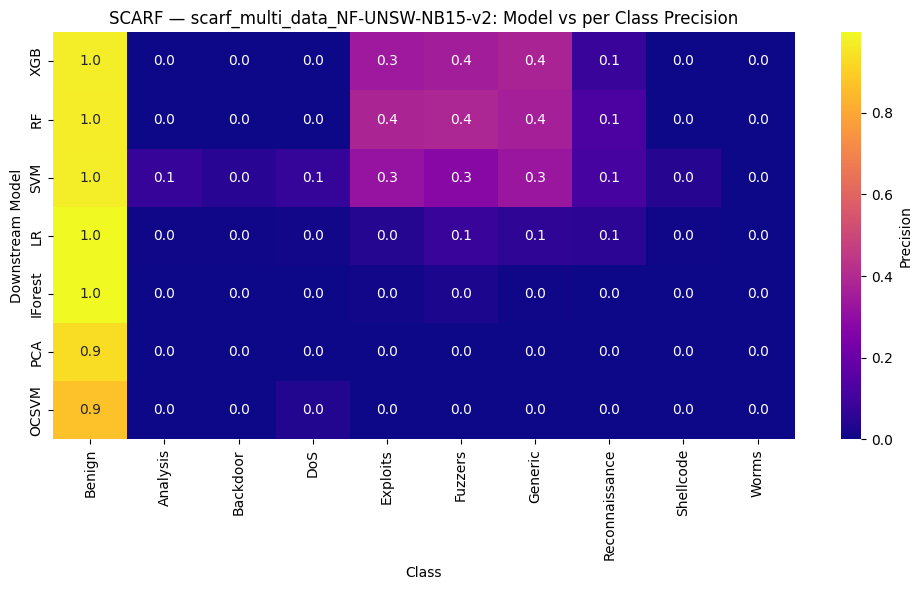}
      \caption{SCARF Precision per class on UNSW-NB15}
    \end{minipage}
    \hfill
    \begin{minipage}{0.45\textwidth}
      \centering
      \includegraphics[width=\linewidth]{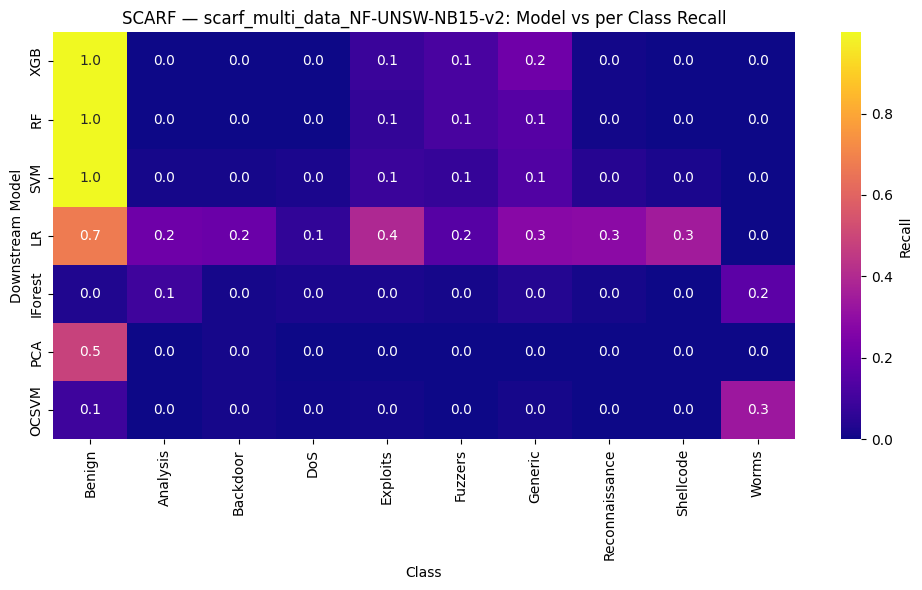}
      \caption{SCARF Recall per class on UNSW-NB15}
    \end{minipage}
    
  \vspace{0.5cm}
  
  \begin{minipage}{0.45\textwidth}
    \centering
    \includegraphics[width=\linewidth]{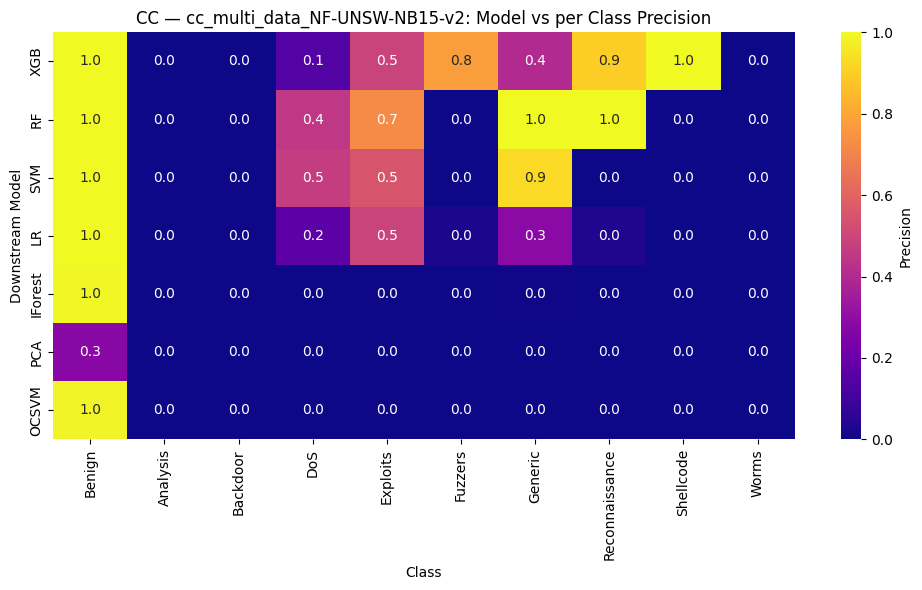}
    \caption{CC Precision per class on UNSW-NB15}
  \end{minipage}
  \hfill
  \begin{minipage}{0.45\textwidth}
    \centering
    \includegraphics[width=\linewidth]{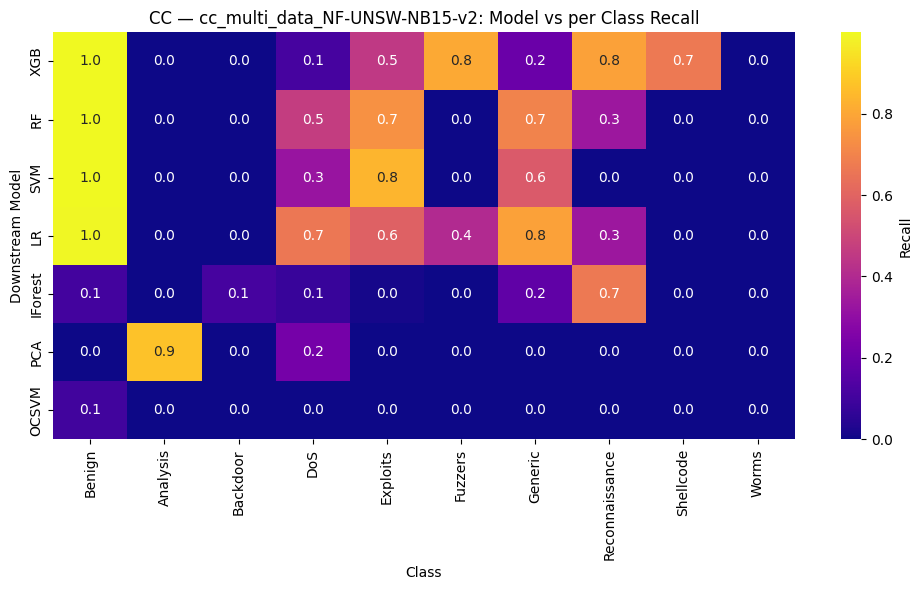}
    \caption{CC Recall per class on UNSW-NB15}
  \end{minipage}
\end{figure}

\clearpage

\emph{CSE-CIC-IDS2018 Dataset: Multiclass Performance}

\begin{figure}[H]
    \centering
  \begin{minipage}{0.45\textwidth}
    \centering
    \includegraphics[width=\linewidth]{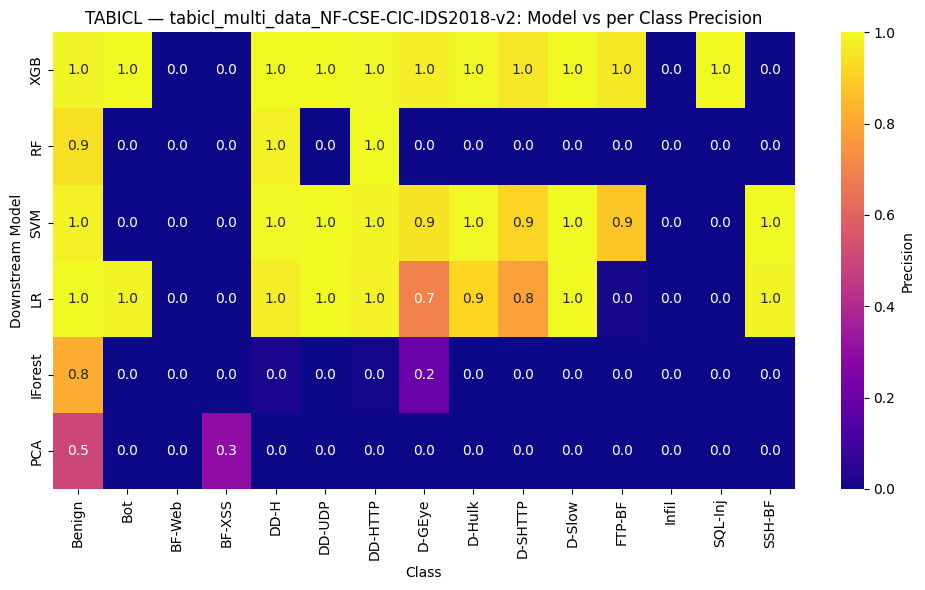}
    \caption{TabICL Precision per class on CSE-CIC-IDS2018}
  \end{minipage}
  \hfill
  \begin{minipage}{0.45\textwidth}
    \centering
    \includegraphics[width=\linewidth]{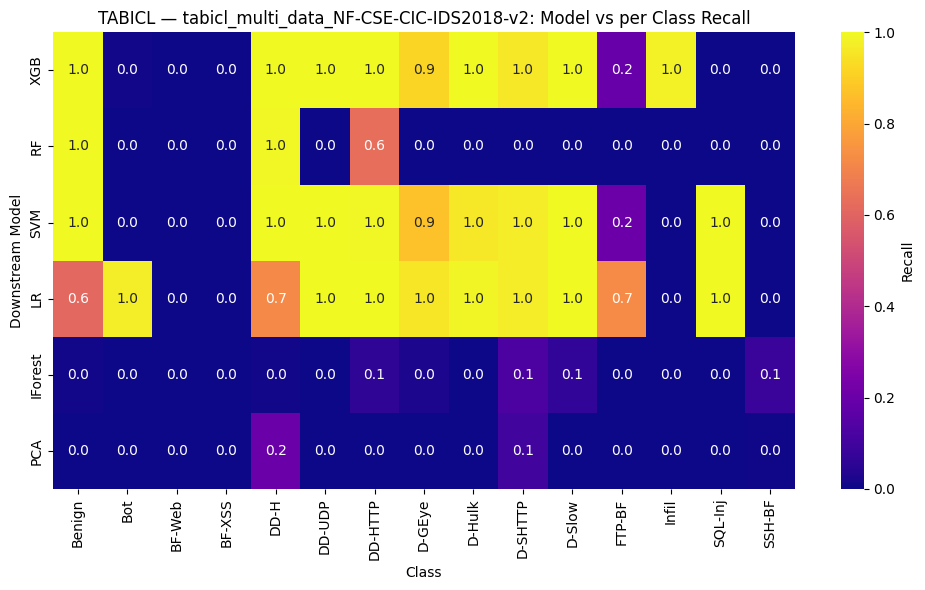}
    \caption{TabICL Recall per class on CSE-CIC-IDS2018}
  \end{minipage}
  
  \vspace{0.5cm}

    \begin{minipage}{0.45\textwidth}
      \centering
      \includegraphics[width=\linewidth]{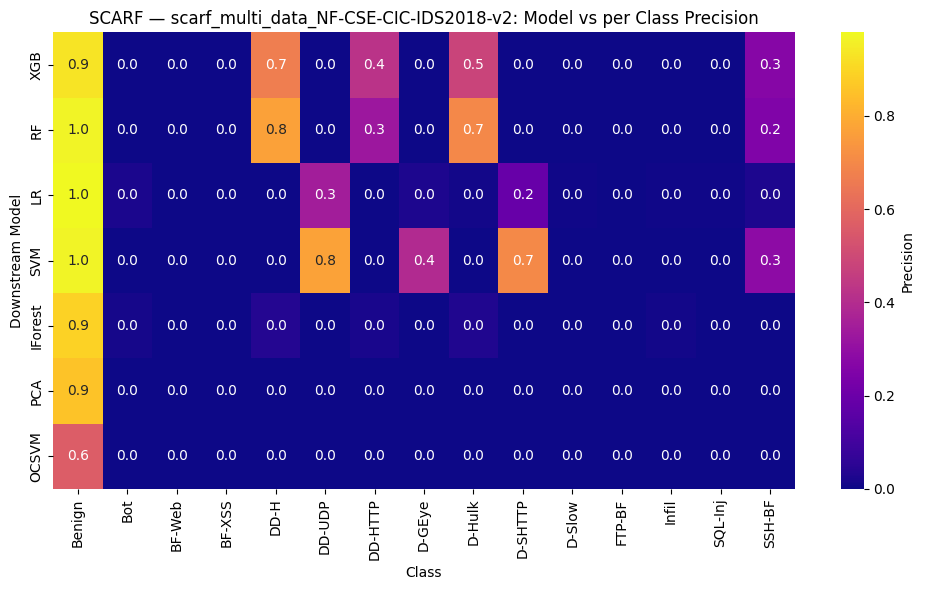}
      \caption{SCARF Precision per class on CSE-CIC-IDS2018}
    \end{minipage}
    \hfill
    \begin{minipage}{0.45\textwidth}
      \centering
      \includegraphics[width=\linewidth]{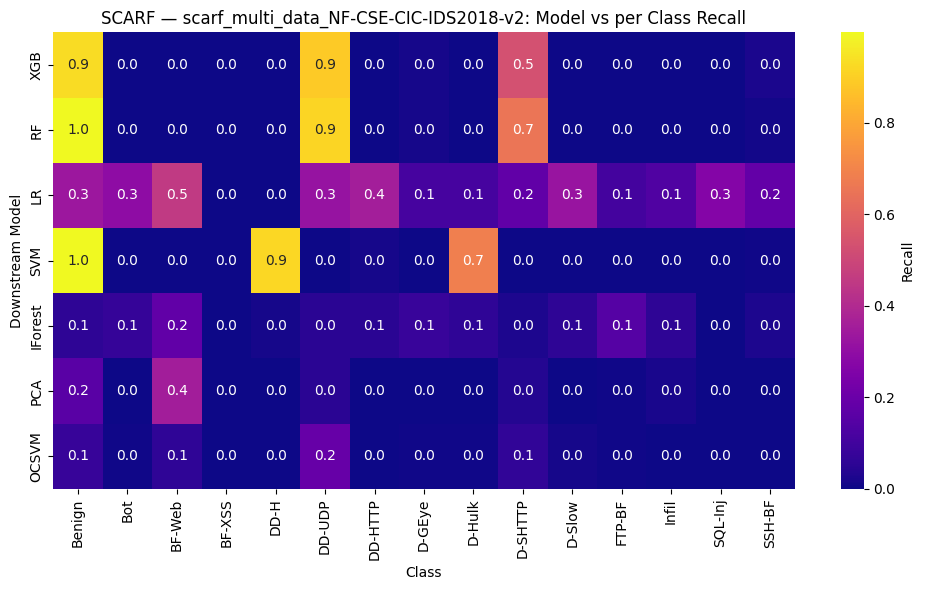}
      \caption{SCARF Recall per class on CSE-CIC-IDS2018}
    \end{minipage}

  \vspace{0.5cm}
  
  \begin{minipage}{0.45\textwidth}
    \centering
    \includegraphics[width=\linewidth]{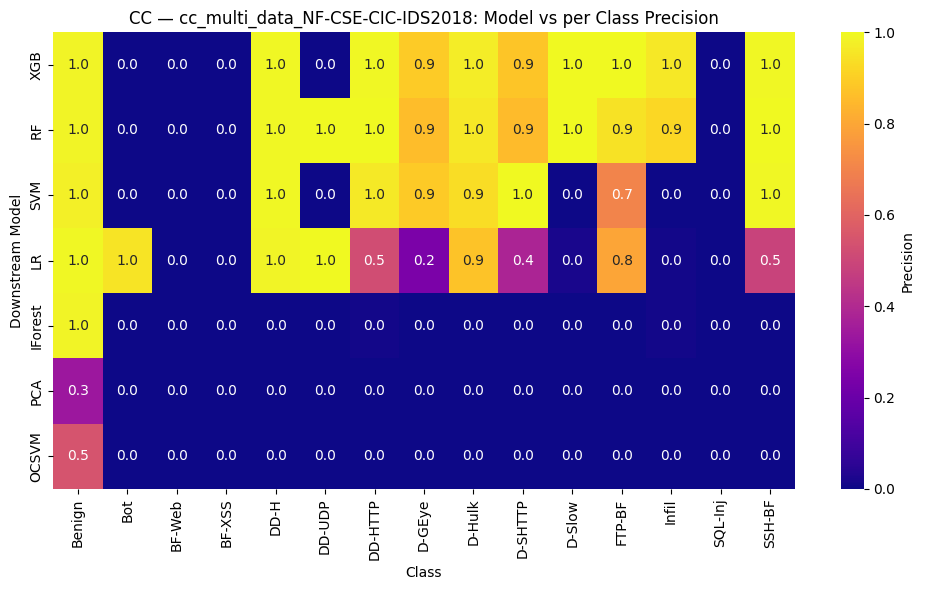}
    \caption{CC Precision per class on CSE-CIC-IDS2018}
  \end{minipage}
  \hfill
  \begin{minipage}{0.45\textwidth}
    \centering
    \includegraphics[width=\linewidth]{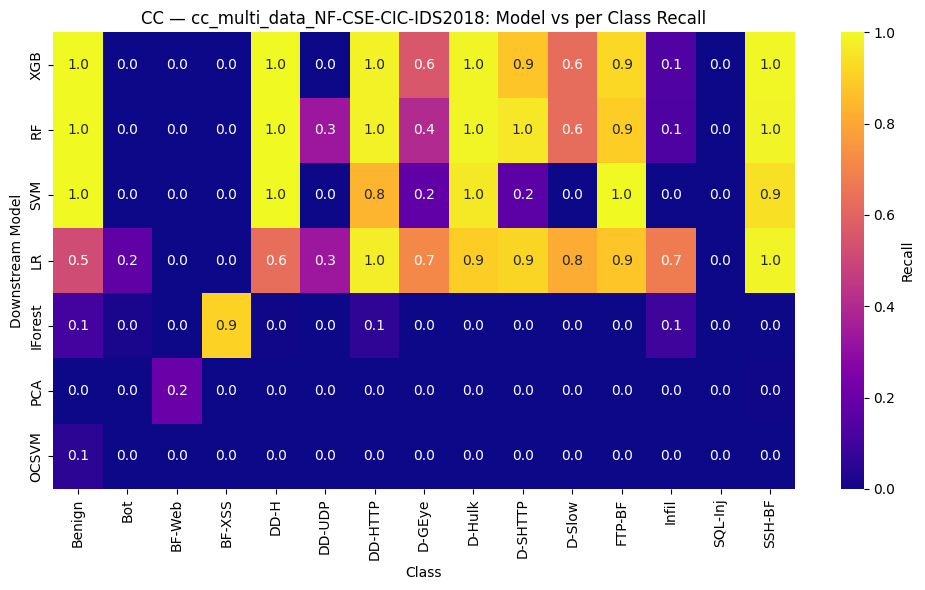}
    \caption{CC Recall per class on CSE-CIC-IDS2018}
  \end{minipage}
\end{figure}

\clearpage

\emph{F1 Score and Macro F1 Analysis}

\begin{figure}[H]
  \centering
  \begin{minipage}{0.45\textwidth}
    \centering
    \includegraphics[width=\linewidth]{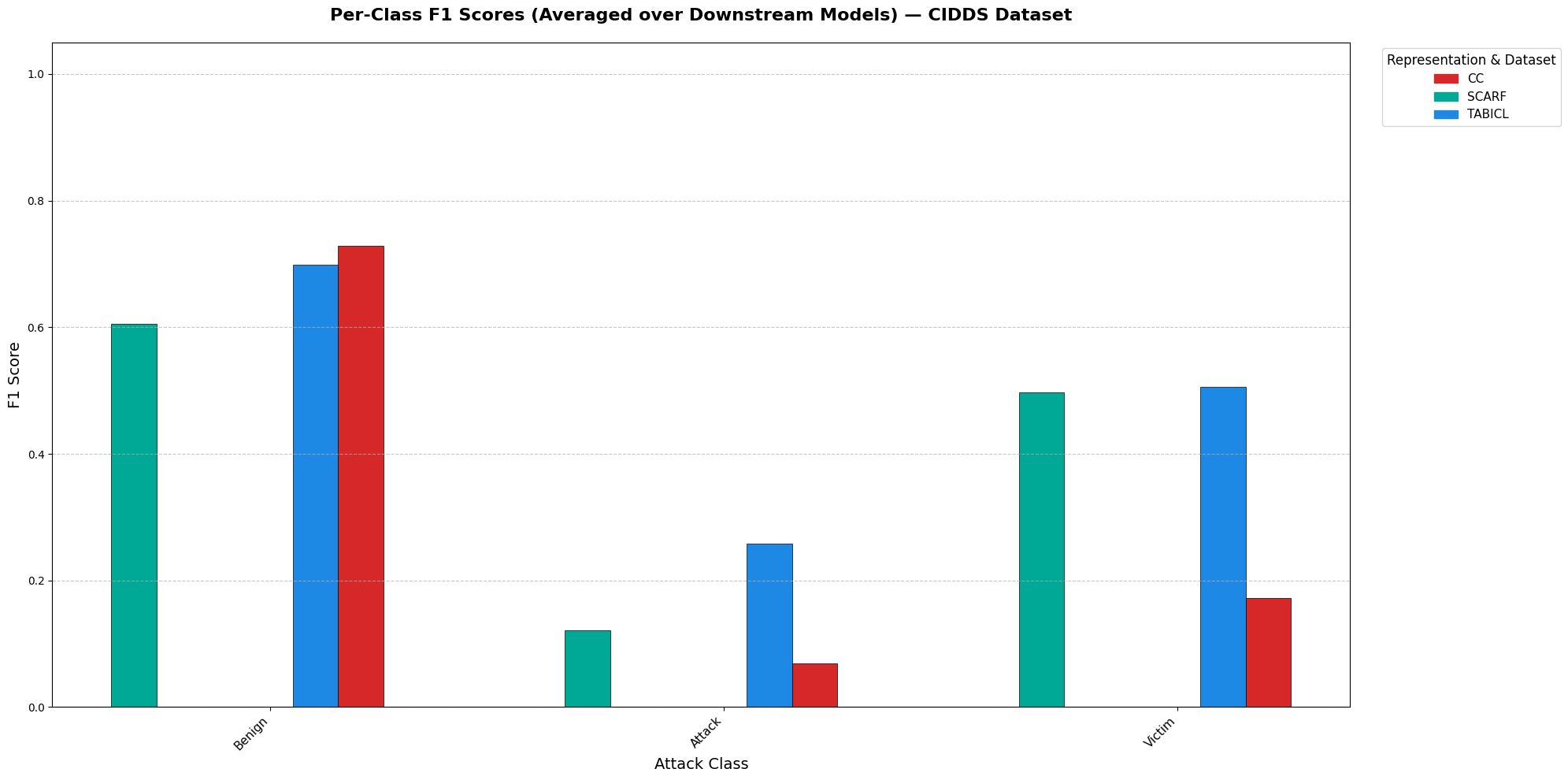}
    \caption{Per Class F1 score on CIDDS-001}
  \end{minipage}
  \hfill
  \begin{minipage}{0.45\textwidth}
    \centering
    \includegraphics[width=\linewidth]{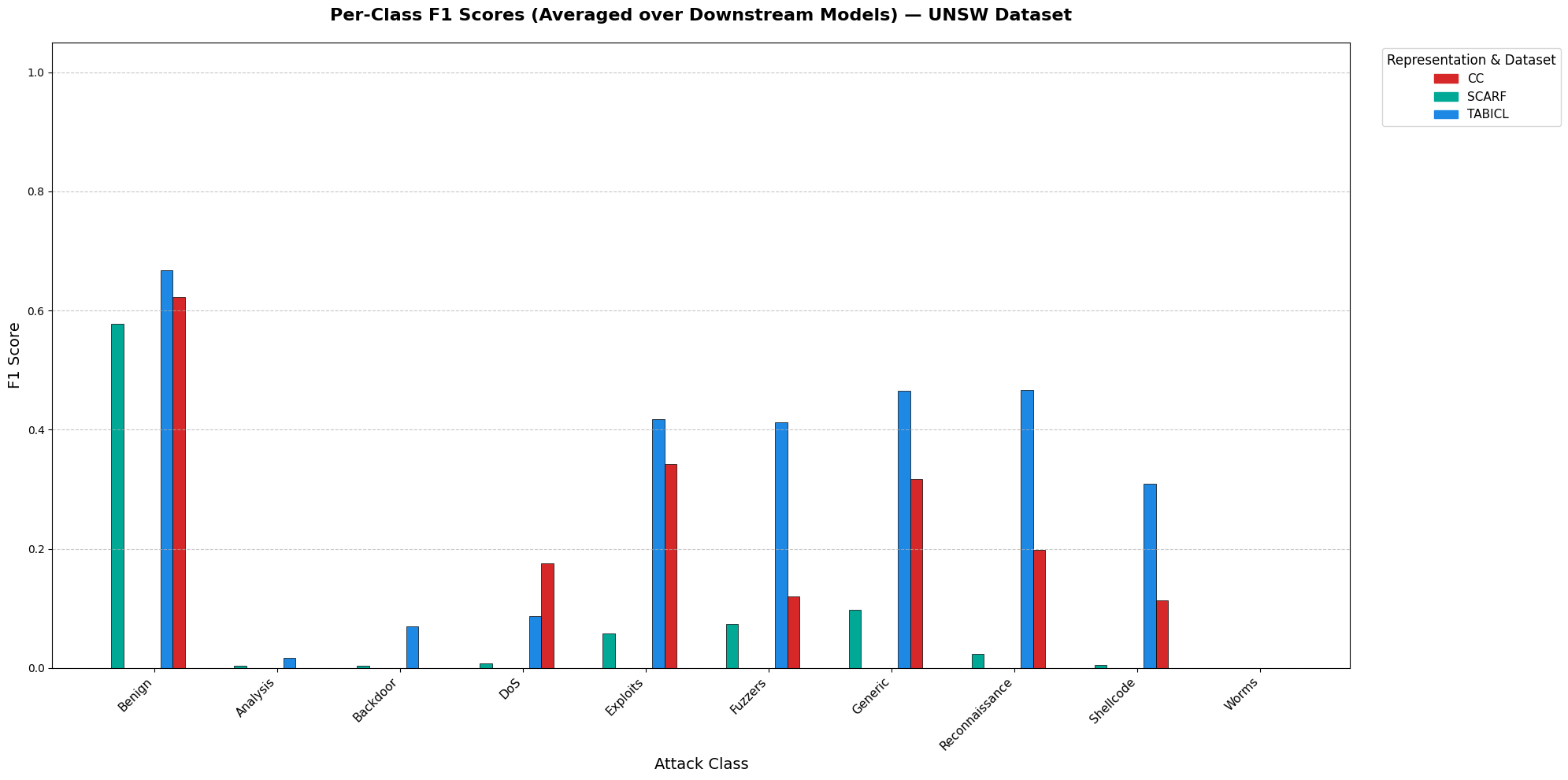}
    \caption{Per Class F1 score on UNSW-NB15}
  \end{minipage}
  
  \vspace{0.5cm}
  
  \begin{minipage}{0.45\textwidth}
    \centering
    \includegraphics[width=\linewidth]{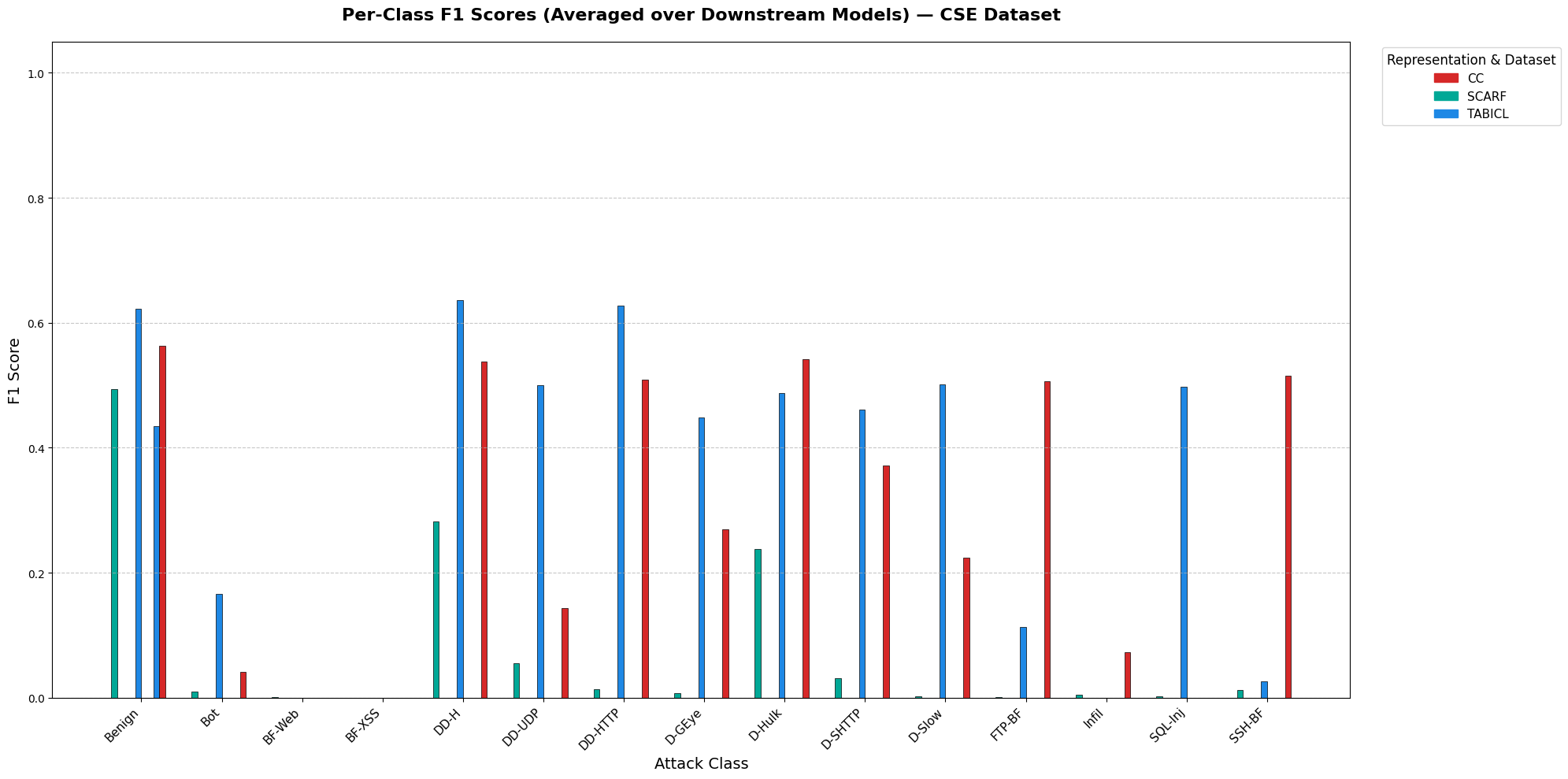}
    \caption{Per Class F1 score on CSE-CIC-IDS2018}
  \end{minipage}
  \hfill
  \begin{minipage}{0.45\textwidth}
    \centering
    \includegraphics[width=\linewidth]{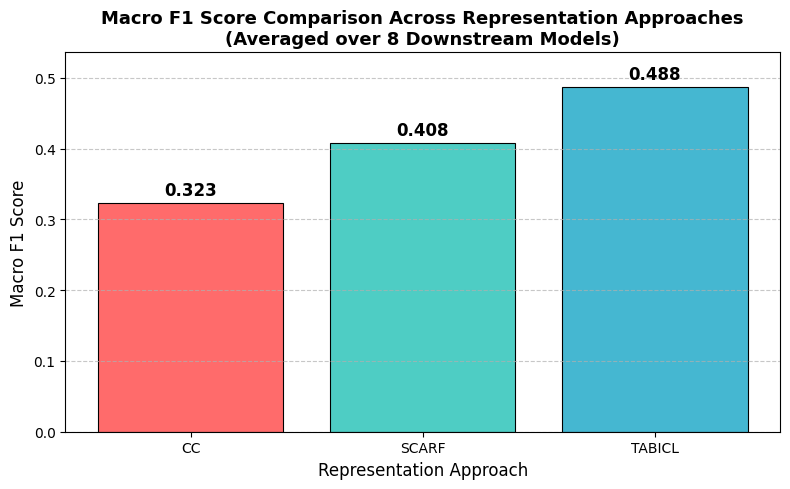}
    \caption{Average Macro F1 on CIDDS-001}
  \end{minipage}
  
  \vspace{0.5cm}
  
  \begin{minipage}{0.45\textwidth}
    \centering
    \includegraphics[width=\linewidth]{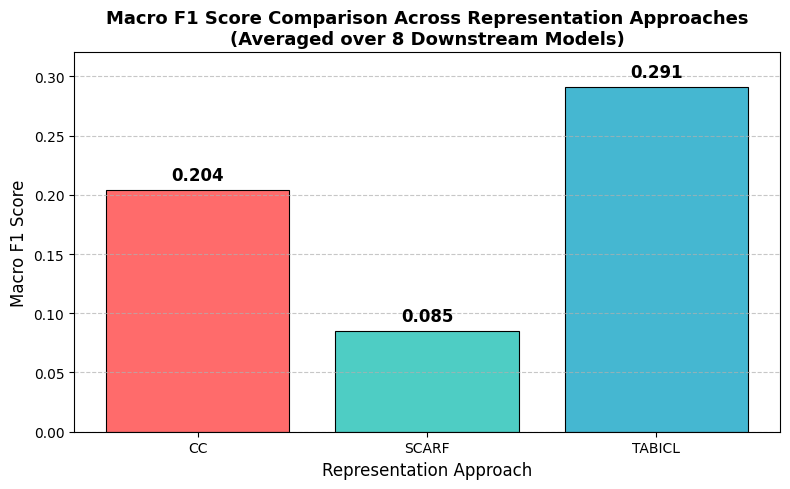}
    \caption{Average Macro F1 on UNSW-NB15}
  \end{minipage}
  \hfill
  \begin{minipage}{0.45\textwidth}
    \centering
    \includegraphics[width=\linewidth]{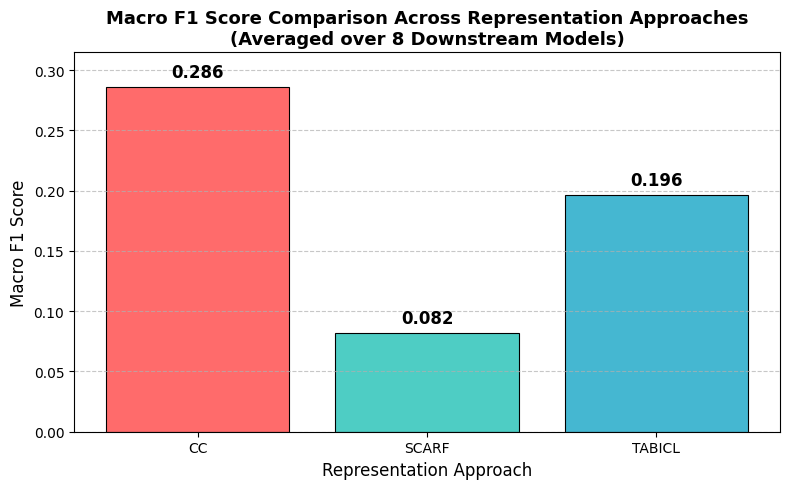}
    \caption{Average Macro F1 on CSE-CIC-IDS2018}
  \end{minipage}
\end{figure}

\end{document}